\documentclass{article}
\usepackage{arxiv}

\usepackage{setspace}
\usepackage{microtype}
\usepackage{graphicx}
\usepackage{subfigure}
\usepackage{booktabs} 
\usepackage{enumitem}
\usepackage{hyperref}
\usepackage{algorithm}
\usepackage{algorithmic}

\usepackage{amsmath}
\usepackage{amssymb}
\usepackage{mathtools}
\usepackage{amsthm}
\usepackage{multirow}

\theoremstyle{plain}

\theoremstyle{definition}

\theoremstyle{remark}

\usepackage[numbers]{natbib}
\newcommand{\methodName}[0]{SALSA-RL}

\usepackage[textsize=tiny]{todonotes}

\title{\methodName{}: Stability Analysis in the Latent Space of Actions for Reinforcement Learning}

\author{ \hspace{1mm}Xuyang Li\\
	College of Information Sciences and Technology\\
	Pennsylvania State University\\
	University Park, PA 16802\\
	\texttt{lixuyang@psu.edu} \\
	\And
	\hspace{1mm}Romit Maulik\\
	College of Information Sciences and Technology\\
	Pennsylvania State University\\
	University Park, PA 16802 \\
	\texttt{rmaulik@psu.edu} \\
}

\renewcommand{\shorttitle}{\textit{arXiv} \methodName{}}

\hypersetup{
pdftitle={\methodName{}},
pdfauthor={Xuyang Li, Romit Maulik},
pdfkeywords={reinforcement learning, dynamical system, stability analysis},
}

\begin{document}
\maketitle
\begin{abstract}
Modern deep reinforcement learning (DRL) methods have made significant advances in handling continuous action spaces. However, real-world control systems, especially those requiring precise and reliable performance, often demand interpretability in the sense of a-priori assessments of agent behavior to identify safe or failure-prone interactions with environments. To address this limitation, this work proposes \methodName{} (Stability Analysis in the Latent Space of Actions), a novel RL framework that models control actions as dynamic, time-dependent variables evolving within a latent space. By employing a pre-trained encoder-decoder and a state-dependent linear system, this approach enables interpretability through local stability analysis, where instantaneous growth in action-norms can be predicted before their execution.
It is demonstrated that \methodName{} can be deployed in a non-invasive manner for assessing the local stability of actions from pretrained RL agents without compromising on performance across diverse benchmark environments. By enabling a more interpretable analysis of action generation, \methodName{} provides a powerful tool for advancing the design, analysis, and theoretical understanding of RL systems.
\end{abstract}

\section{Introduction}
Reinforcement learning (RL) is a powerful framework for training agents to make sequential decisions directly from environment interactions \cite{sutton2018reinforcement}, and it has shown remarkable success in complex continuous control tasks \cite{lillicrap2015continuous}. Unlike discrete decision-making tasks (e.g., Chess or Go), real-world dynamical systems evolve continuously, often requiring fine-grained and highly accurate control inputs to ensure safe and robust operation. In these settings, even small errors in control can propagate quickly, leading to instability, unpredictable behavior, and potential system failures. Despite these risks, most existing DRL approaches \cite{yu2023reinforcement, heuillet2021explainability, kalashnikov2018scalable} do not provide explicit mechanisms for analyzing or ensuring stability, an omission that poses significant challenges in safety-critical domains.

In dynamical system control, actions play a central role as they directly influence state evolution by serving as the primary inputs driving changes in the state trajectory. While state trajectories reflect the physical behavior of the system, they often fail to reveal the reasoning behind the control decisions. Particularly, different policies can produce similar state trajectories while relying on fundamentally different decision patterns, making state-based observations insufficient for fully understanding control logic.
In contrast, action dynamics, which describe how control inputs evolve over time in response to system feedback, offer a more concrete way to assess the structure of a control policy. By analyzing trends and variations in action sequences, critical patterns can emerge, such as smooth and consistent adjustments indicating stable regulation, abrupt shifts reflecting reactive strategies, or periodic oscillations suggesting transience even in stable regimes.
In the LunarLander problem \cite{brockman2016openai}, for instance, multiple strategies can achieve a successful landing, but the specific thrust and rotation sequences can differ greatly, indicating distinct approaches to stabilization and error correction.

Additionally, observing how actions evolve provides insight into a policy's ability to maintain stability by avoiding erratic decision patterns, adapting to disturbances with smooth control adjustments, and balancing exploration and exploitation for effective regulation without excessive correction. This analysis also enhances RL interpretability through an answer to the following questions:

\begin{enumerate}[nosep]
\item Given pretrained reinforcement learning agents, {\color{black}is it possible to} identify regions of smooth or regular agent behavior in action-state phase space?
\item Conversely, can regions with a high degree of discontinuous (and consequently high-risk/failure-prone) behavior {\color{black}also be identified}?
\end{enumerate}

Identifying such patterns not only aids in policy validation but also helps detect subtle issues like overcompensation, delayed reactions, or unstable oscillations that may not be evident when observing state or action trajectories in their original high-dimensional spaces. {\color{black} Moreover, in many physical systems, actions are often the only fully observable or controllable signals, whereas the state may be difficult or impractical to measure directly. In addition, actions are typically subject to physical constraints or actuation budgets that may not be fully embedded during learning. Anticipating how actions behave, especially under changing conditions, can provide valuable insights for safety and planning.}

In light of the importance of action dynamics for understanding control strategies, \methodName{} is proposed, a novel RL framework for dynamical system control that leverages action dynamics to analyze and design stable control strategies while enhancing interpretability. 
Unlike standard RL approaches, where actions are treated as discrete decisions, this framework models actions in a latent space, evolving based on the state information.
Specifically, the framework employs a latent representation governed by a state-dependent linear dynamical system.
This allows for local stability analysis of action dynamics in the latent space, providing information about the short-term growth of action-dynamic norms.
{\color{black}
Crucially, this local stability metric serves as a diagnostic proxy for the internal consistency and regularity of the control policy, rather than a formal mathematical guarantee of closed-loop physical safety or an absolute indicator of system failure.}
Moreover, such structured representation not only allows for local-stability analysis of control policies but also provides a means to visualize and interpret the constraints imposed on the state space, enhancing both policy transparency and generalization. 

This proposed framework is compatible with existing popular DRL methods (such as SAC, DDPG, TD3 \cite{haarnoja2018soft, fujimoto2018addressing, lillicrap2015continuous}) and can be easily integrated using exactly the same standard training strategies while maintaining competitive control performance. 
Beyond achieving effective control, \methodName{} provides valuable tools for evaluating trained policies. 
{\color{black}Since actions directly influence state evolution, the local stability of action dynamics serves as an indirect yet effective indicator of state trajectory stability.
Furthermore, incorporating local-in-time eigenvalue-based stability analysis enables a rigorous evaluation of the system's behavior.
This provides insights into how action-space constraints impact long-term stability and how state transitions behave near critical regions, further enhancing the interpretability of the learned control policy.}

The primary contributions are summarized as follows:
\begin{enumerate}[nosep]
\item This work introduces \methodName{}, a framework that models control actions as discrete-time dynamics in latent space via a time-varying linear system, enabling deeper insight into control strategies and dynamics.
\item \methodName{} integrates seamlessly with standard DRL methods, functioning as a non-invasive, post-hoc diagnostic tool that enhances interpretability without compromising the control performance.
\item Local stability analysis is performed to identify regions of action-state phase space for regular and high-risk behavior often undetectable through standard state-trajectory observations. It is also demonstrated that pretrained agents seek out regions of regularity.
\end{enumerate}

{\color{black}The remainder of this paper is organized as follows. Section 2 reviews related work in classical control, interpretability in deep reinforcement learning, and dynamical system control. Section 3 details the proposed \methodName{} framework, including its problem formulation, latent dynamic modeling, and numerical techniques for stability analysis. Section 4 presents the experimental evaluation across various continuous control benchmarks, along with an ablation study. Finally, Section 5 concludes the paper and outlines potential directions for future research.}

\section{Related Work}
\subsection{Classical Control} 
Classical approaches for dynamical system controls include Proportional-Integral-Derivative (PID), Linear Quadratic Regulators (LQRs), and adaptive control techniques\cite{aastrom2006advanced, swarnkar2014adaptive}. These methods offer clear advantages in interpretability and stability analysis by providing explicit symbolic expressions for the relationship between states and control actions, enabling stability analysis and control law verification. However, they often struggle with high-dimensional, nonlinear, or partially observable systems where dynamics cannot be explicitly modeled \cite{skogestad2005multivariable}. {\color{black}\methodName{} bypasses explicit physical modeling by evaluating stability directly on learned latent action dynamics.}

\subsection{Interpretability in DRL} 
Interpretability for machine learning algorithms has gained significant attention in recent years, particularly in healthcare, robotics, and autonomous systems \cite{glanois2024survey, yu2023reinforcement, heuillet2021explainability}. Recent research devoted to interpretable RL has explored diverse approaches such as multi-agent systems \cite{zabounidis2023concept}, interpretable latent representations \cite{chen2021interpretable}, and techniques like attention mechanisms \cite{mott2019towards} or genetic programming \cite{hein2018interpretable}, balancing transparency, performance, and generalizability. {\color{black}Recent studies have also highlighted the importance of efficient interpretability and adaptability in RL through techniques such as policy distillation \cite{xing2023achieving}, the extraction of explainable, task-relevant state representations \cite{zhao2024learning}, and unsupervised exploration via self-referencing mechanisms \cite{zhao2025self}.} Furthermore, while embedding time-dependent dynamics into pre-trained models enhances static classification \cite{ding2025learning}, \methodName{} distinctly applies this concept to continuous action spaces to evaluate physical stability in sequential decision-making.
Several key directions are outlined below.

\textbf{Hierarchical RL.} This approach focuses on planning and reasoning by structuring tasks into high-level (manager) and low-level (worker) policies \cite{lyu2019sdrl, nachum2018data, pateria2021hierarchical}. This modular approach excels in task decomposition, reusable subtasks, and explainable goal-setting. However, it is less suitable for specific use cases like dynamical system control, where precise and responsive low-level policies are essential \cite{florensa2017stochastic}. 

\textbf{Prototype-based RL.} Prototype-based methods leverage human-defined prototypes to represent states and actions as interpretable latent features, enabling policies that balance interpretability and performance \cite{kenny2023towards, xiong2023interpretable, yarats2021reinforcement}. However, their reliance on manual design and lack of temporal precision limit their adaptability to dynamic environments and their ability to handle continuous, high-dimensional dynamics effectively \cite{duan2016benchmarking}. {\color{black}To address temporal imprecision and manual design limits, \methodName{} dynamically extracts stability metrics from continuous action sequences.}

\subsection{Dynamical System Control and Deep Reinforcement Learning}
Methods such as Koopman-based control~\cite{lyu2023task, yeung2019learning} and Embed-to-Control~\cite{watter2015embed} model nonlinear dynamics through latent linearization, {\color{black}where the environment state is approximated as linear in a lifted latent space for system identification}. 
Though not inherently RL, these methods have inspired RL frameworks with structured latent dynamics that often rely on globally fixed linear operators, rather than state-dependent ones, thereby limiting adaptability to state-dependent variations in very high-dimensional nonlinear dynamical systems~\cite{weissenbacher2022koopman, rozwood2024koopman}. In particular, \cite{rozwood2024koopman} introduces value function learning with dictionary-based methods that can provide increased insight into agent behavior. However, a direct connection to the inherent local regularity of the trained agent's behavior is not clear.
{\color{black}Fundamentally distinct from these Koopman-based approaches, \methodName{} does not attempt to model the physical environment's state dynamics. Instead, it lifts the agent's actions into a latent space governed by a state-dependent, time-varying dynamic matrix. This shift from global state linearization to local action dynamics modeling enables explicit stability analysis of the policy's internal decision-making process.}

On the other hand, DRL is well-studied in physical applications, such as optimizing flow control and turbulent fluid dynamics, where domain knowledge is often available, and state-control actions exhibit stronger correlations compared to other domains~\cite{yousif2023physics, weiner2024model, garnier2021review}. 
{\color{black}Recent advancements extend DRL to complex distributed physical systems, including continuous resource allocation with rigorous error control \cite{wang2024joint, zhang2024unified} and highly dynamic maritime operations involving multi-USV path planning, collision avoidance, and real-time decision-making \cite{yin2025adaptive, chen2025dynamic, yin2025real}. The deployment in such safety-critical applications underscores the strict necessity for reliable and interpretable control policies.}
Recent approaches (summarized below) have further improved interpretability by leveraging state-action correlations and domain knowledge.

\textbf{Symbolic and Neural-Guided Approaches} 
Neural-guided methods like NUDGE \cite{delfosse2024interpretable}, PIRL \cite{verma2018programmatically}, and NLRL \cite{jiang2019neural} use symbolic reasoning to discover interpretable policies \cite{jin2022creativity}. Symbolic controllers \cite{khaled2022framework, reissig2018symbolic}, such as DSP \cite{landajuela2021discovering} and SINDy-RL \cite{zolman2024sindy}, employ explicit state-control mappings to derive generalizable control laws.
These methods improve interpretability and generalizability, making them suitable for safe, explainable decision-making. However, neural-guided approaches rely on logical representations \cite{delfosse2024interpretable}, limiting temporal precision and scalability in complex systems. Similarly, symbolic controllers face scalability issues in multi-dimensional or stochastic tasks where predefined forms fail to capture intricate interactions \cite{landajuela2021discovering}. While explicit formulations may aid analysis of agent behavior, explicit stability assessments for complex high-dimensional learning tasks are challenging. {\color{black}Instead of relying on rigid symbolic formulations, \methodName{} utilizes state-dependent linear representations for scalable stability evaluations.}

\textbf{Physics-guided RL}
Physics-guided RL \cite{liu2021physics, banerjee2023survey, alam2021physics, wang2022toward} integrates domain knowledge and physical principles to enhance learning and performance. Approaches include informed reward functions, model-based RL with physics-based or neural network surrogate models \cite{hernandez2023port, nousiainen2021adaptive}, and state design \cite{banerjee2023survey}. While incorporating domain knowledge can improve interpretability, these methods \cite{alam2021physics, jurj2021increasing, cho2019physics, wang2022toward} often lack generalizability to other problems and require significant fine-tuning. {\color{black}As an algorithm-agnostic framework, \methodName{} ensures broader applicability without requiring problem-specific physical equations or extensive fine-tuning.}

\section{Method}
{\color{black}{\color{black}Notation: Throughout this paper, scalars are denoted by standard italic letters (e.g., $t, \gamma, r$), vectors by bold lowercase letters (e.g., $\mathbf{s}_t, \mathbf{a}_t, \mathbf{z}_t$), and matrices by bold uppercase letters (e.g., $\mathbf{A}_t, \boldsymbol{\Phi}_t$). Continuous spaces are denoted by calligraphic uppercase letters (e.g., $\mathcal{S}, \mathcal{A}$). Standard reinforcement learning functions (e.g., action-value function $Q$, transition probability $P$) follow conventional notations, while specific neural network mappings are represented by cursive or Greek letters (e.g., $\mathcal{N}_\theta, \pi_\theta$). Note that $\theta$ denotes neural network parameters generally, except in Section 4 where it conventionally represents the pendulum angle.}}

\subsection{Problem Formulation}
Consider a control system with continuous dynamics,
\begin{equation}
  \dot{\mathbf{s}}(t) \;=\; \mathcal{F}\bigl(\mathbf{s}(t), \mathbf{a}(t)\bigr),
\end{equation}
where $\mathbf{s}(t) \in \mathcal{S} \subseteq \mathbb{R}^n$ is the state and $\mathbf{a}(t) \in \mathcal{A} \subseteq \mathbb{R}^m$ is the control input. Discretizing the inputs at intervals $\Delta t$ yields a discrete-time Markov Decision Process (MDP) $\mathcal{M} = (\mathcal{S}, \mathcal{A}, P, R, \gamma)$, where $\mathcal{S}$ represents the state space and $\mathcal{A}$ the action space governing the system's behavior.

Assuming the Markov property, the next state $\mathbf{s}_{t+1}$ depends on the current state $\mathbf{s}_t$ and action $\mathbf{a}_t$, governed by the transition probability $P(\mathbf{s}_{t+1} \mid \mathbf{s}_t, \mathbf{a}_t)$. The agent maximizes the cumulative discounted reward $R = \sum_{t=0}^{T-1} \gamma^t r(\mathbf{s}_t, \mathbf{a}_t)$, where $\gamma \in [0,1]$ balances immediate and future rewards over the time horizon $T$.

\begin{figure}[ht!]
\begin{center}
\centerline{\includegraphics[width=0.8\columnwidth]{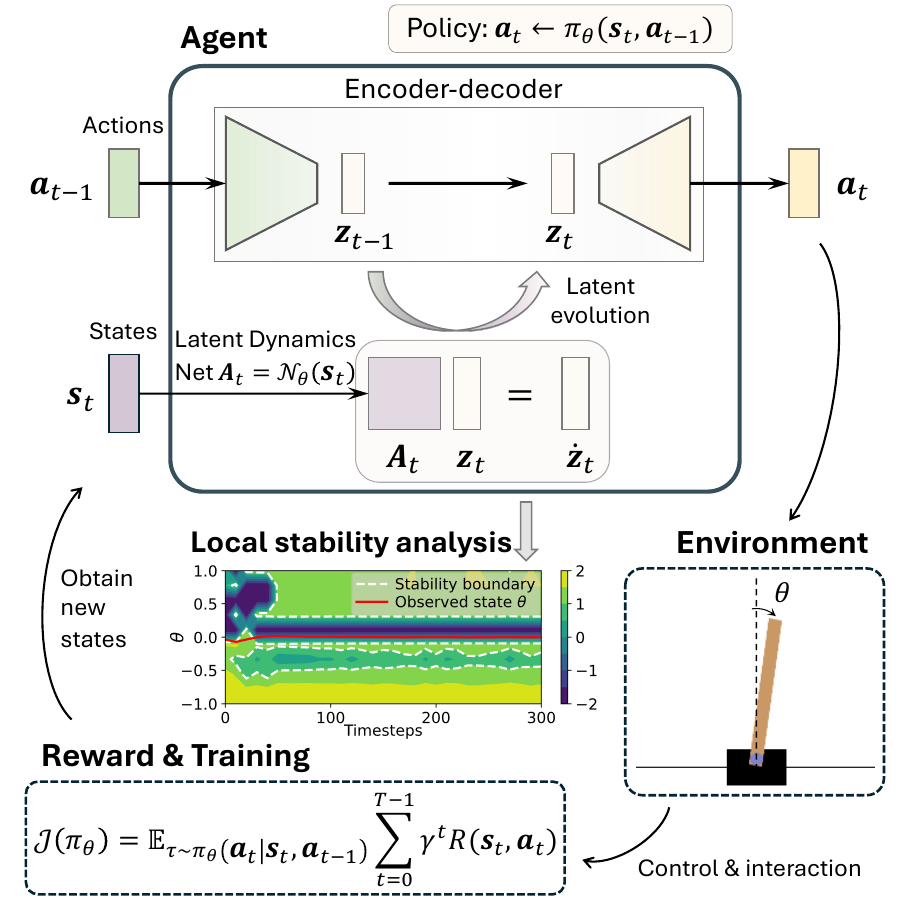}}
\caption{\textbf{Overview} of the \methodName{} framework. The architecture augments pre-trained RL agents by modeling control actions within a latent space governed by a state-conditioned linear dynamical system, $\mathbf{A}_t$. The contours visualize the time-varying spectral radius $\rho(\mathbf{A}_t)$, serving as a metric for local stability analysis that identifies stable and unstable regions within the action-state phase space.}
\label{fig:overview}
\end{center}
\end{figure}

Unlike standard DRL, which utilizes only state-dependent policies $\pi_\theta(\mathbf{a}_t \mid \mathbf{s}_t)$, \methodName{} relies on actions in a latent space generated by a dynamical system, resulting in the policy $\pi_\theta(\mathbf{a}_t \mid \mathbf{s}_t, \mathbf{a}_{t-1})$. The policy is optimized to maximize the expected cumulative return:
\begin{equation}
  J(\pi_\theta)
  \;=\;
  \mathbb{E}_{\tau \sim \pi_\theta(\mathbf{a}_t \mid \mathbf{s}_t, \mathbf{a}_{t-1})}
  \Bigl[
    \sum_{t=0}^{T-1} \gamma^t \, R\bigl(\mathbf{s}_t, \mathbf{a}_t\bigr)
  \Bigr],
\end{equation}
where $\tau= (\mathbf{s}_0, \mathbf{a}_0, \dots, \mathbf{s}_{T}, \mathbf{a}_{T})$ represent trajectories sampled from the policy. Although the latent action $\mathbf{z}$ evolves in continuous time, the discrete-time MDP formulation is sufficient for the proposed algorithm, as actions are applied at discrete intervals $\Delta t$. Figure \ref{fig:overview} illustrates the details of \methodName{}.

\subsection{\methodName{} Control Framework\label{sec:method}}
The proposed framework simplifies control of time-dependent dynamical systems by encoding actions into a compact latent space, where state-dependent transformations govern their evolution. It consists of:

\textbf{Action Encoding and Latent Representation.} The policy's action $\mathbf{a}_t$ is encoded into a latent space using a pre-trained autoencoder:
\begin{equation} \label{eq:encoder}
   \mathbf{z}_t = \text{Encoder}(\mathbf{a}_t),
\end{equation}
where $\mathbf{z}_t \in \mathbb{R}^{h_d}$ is the latent representation of the action, $h_d$ is the latent dimension size, and $\mathbf{a}_0=\mathbf{0}$. Here, actions are intentionally projected into a higher-dimensional latent space (typically $h_d > \dim(\mathbf{a})$) to provide greater expressive capacity for learning structured dynamics. This higher-dimensional representation eases training and supports more interpretable stability analysis, especially in complex or underactuated tasks.
{\color{black} Crucially, a standard deterministic autoencoder is employed rather than a Variational Autoencoder (VAE). The stochastic sampling inherent to VAEs would inject continuous noise into the decoded actions, corrupting the absolute precision required for closed-loop dynamic control and artificially distorting the eigenvalues of the stability analysis.}

\textbf{Latent Dynamics Module.} The latent action representation evolves according to a learned state-dependent linear dynamical system $\dot{\mathbf{z}}_t = \mathbf{A}_t \mathbf{z}_t$. Simply, for discrete actions:  
\begin{align}
   \mathbf{z}_{t+1} &= \mathbf{z}_t+\mathbf{A}_t\mathbf{z}_t \label{eq:Az}, \\
   \mathbf{A}_t &= \mathcal{N}_\theta(\mathbf{s}_t), \label{eq:fs}
\end{align}
where $\mathbf{A}_t \in \mathbb{R}^{h_d \times h_d}$ is a state-dependent matrix learned by the neural network $\mathcal{N}_\theta$ conditioned on the current state. Importantly, to ensure numerical stability, a $\tanh$ activation function is applied, followed by a scaling factor of 2 to the output of the latent dynamics network, thereby constraining all elements of $\mathbf{A}_t$ within the range $[-2, 2]$. While diagonal forms of $\mathbf{A}_t$ were also explored to simplify the analysis, full (dense) matrices were found to enable richer inter-dependencies and better performance in control.

{\color{black}It is worth noting that a latent action space is introduced not for dimensionality reduction but to enable structured analysis of action behavior in the state–action phase space. This design empirically yields smoother dynamics and more stable training than operating directly in the original action space, while also allowing flexible dimensionality for constructing square dynamics matrices. It further aligns with Koopman operator theory, where systems are lifted to higher-dimensional spaces for structured modeling before being mapped back.}

\textbf{Action Decoding and Control Execution.} After the linear system evolution, the latent representation is decoded back into the original action space as the actual control policy:  
\begin{equation}
   \mathbf{a}_{t+1} = \text{Decoder}(\mathbf{z}_{t+1}). \label{eq:decoder}
\end{equation}
In practice, $\mathbf{z}_{t+1}$ is decoded into an action $\mathbf{a}_{t+1}$ for system control. Then, the same $\mathbf{a}_{t+1}$ is encoded back into a new latent vector $\mathbf{z}_{t+1}'$. This mimics real-world control scenarios where only the executed action is observable, and the internal latent representation must be reconstructed based on the current state and action. This stepwise decode-encode process also reflects how a deployed agent operates under partial observability or noisy dynamics, maintaining consistency through latent transitions.
{\color{black} Additionally, in line with standard RL environment setups, the decoded action is clipped in all experiments. Importantly, this clipping is applied only after decoding; the latent space itself remains unconstrained.}

\textbf{Policy Update and Control Loop.} The system evolves iteratively, where actions generated from the latent space update both the system state and latent variables. The framework is summarized in Algorithm~\ref{alg:1}.

\begin{algorithm}[ht]
   \caption{Latent Dynamic Control Framework}
   \label{alg:1}
\begin{algorithmic}
   \STATE {\bfseries Input:} Initial state $\mathbf{s}_0$, initial action $\mathbf{a}_{-1}=\mathbf{0}$, time horizon $T$, pre-trained \textbf{Encoder} and \textbf{Decoder}
   \FOR{$t = 0$ {\bfseries to} $T$}
       \STATE $\mathbf{z}_t \gets \textbf{Encoder}(\mathbf{a}_{t-1})$ // Encode action
      \STATE $\mathbf{A}_t \gets \mathcal{N}_\theta(\mathbf{s}_t)$
      \STATE $\mathbf{z}_{t+1} \gets \mathbf{z}_t + \mathbf{A}_t \mathbf{z}_t$ // Update latent space
      \STATE $\mathbf{a}_{t} \gets \textbf{Decoder}(\mathbf{z}_{t+1})$ // Decode latent to next action
      \STATE $\mathbf{s}_{t+1}, Done \gets \text{env.step}(\mathbf{a}_{t})$
      \STATE $\mathbf{a}_{t-1} \gets \mathbf{a}_{t}$
      \IF {$Done$}
      \STATE {\bfseries break}
      \ENDIF
   \ENDFOR
\end{algorithmic}
\end{algorithm}

\subsection{Training Procedure}
Training involves two stages: (1) encoder-decoder pretraining, and (2) latent dynamics optimization.

\textbf{Encoder-Decoder Network.} This network is trained to create a compact and effective latent representation of actions, optimized by minimizing the reconstruction loss:
\begin{equation}
    \mathcal{L}_{\text{recon}} = \mathbb{E}_{\mathbf{a}_t \sim D}\left[ \| \mathbf{a}_t - \hat{\mathbf{a}}_t \|^2 \right],
\end{equation}
where $\hat{\mathbf{a}}_t$ is the reconstructed action from the decoder.
{\color{black}To train the module, training data $D$, actions, are constructed from actions drawn either uniformly over the full action space or collected from pretrained agent.}
This data is then used to train the action encoding and decoding modules. This procedure is detailed in~\ref{appendix:encoder-decoder}.

\textbf{Latent Dynamic Policy.} Following this, the proposed policy is initialized in a similar manner as most DRL methods for continuous action control, such as PPO \cite{schulman2017proximal}, SAC, DDPG, and TD3. The policy network (actor), typically generating actions in DRL as $\mathbf{a}_t = \mathcal{N}_\theta(\mathbf{s}_t)$, is instead employed to predict a state-dependent dynamic matrix $\mathbf{A}_t = \mathcal{N}_\theta(\mathbf{s}_t, \mathbf{a}_{t-1})$. The gradients of the objective to maximize the expected cumulative reward, in deterministic policies, for example, can be expressed generally as:
\begin{equation}\label{eq:obj_refined}
\hspace{-1.0mm}
\nabla_\theta J(\pi_\theta)
\;=\;
\mathbb{E}_{\tau \sim \pi_\theta(\mathbf{a}_t \mid \mathbf{s}_t, \mathbf{a}_{t-1})}
\left[
\sum_{t=0}^{T-1} \gamma^t \nabla_\theta R\bigl(\mathbf{s}_t, \mathbf{a}_t\bigr)
\right]
\end{equation}
Meanwhile, the critic network architecture, if available, remains unchanged, evaluating the state-action value function. This design allows \methodName{} to leverage existing DRL algorithms while providing flexibility to represent and optimize latent dynamics. A detailed gradient computation is provided in~\ref{appendix:derivation}.

\subsection{Stability Analysis and Interpretability}

This section explores a range of numerical techniques for analyzing latent action dynamics, comprising the bulk of this work's contribution. {\color{black}To avoid any conceptual ambiguity, it is crucial to first explicitly distinguish between open-loop and closed-loop stability in this context. The following methods evaluate the \textit{open-loop (local) stability} of the policy's internal latent action dynamics, serving as a diagnostic proxy for control regularity. This is fundamentally distinct from the \textit{closed-loop (global) stability} of the physical system, which involves the coupled interactions between the agent's actions and the environment's actual transition dynamics.} 

This stability analysis provides insights into the system's robustness and enhances interpretability, improving understanding of the underlying agent's behavior.
{\color{black} Before detailing the specific techniques, it is essential to distinguish the utility of the latent operator $\mathbf{A}_t$ from simple observational metrics, such as discrete action difference norms ($|| \mathbf{a}_t - \mathbf{a}_{t-1} ||$). While difference norms can retroactively indicate erratic behavior along a realized trajectory, they present fundamental limitations for rigorous policy analysis. Specifically, scalar differences obscure the cross-dimensional couplings inherent in multi-dimensional action spaces and lack the predictive capacity to assess unexplored regions of the state space. 
By modeling control actions through a state-conditioned dynamic operator $\mathbf{A}_t$, \methodName{} addresses these limitations. This formulation preserves multidimensional control interactions and transitions the assessment from an a posteriori observation to an a priori diagnostic tool. Consequently, it unlocks the advanced mathematical evaluations introduced in the subsequent subsections: local eigenvalue analysis, transient growth, and Floquet stability.}

\textbf{Local Stability Analysis.} While the environments are continuous-time, the action dynamics modeled in \methodName{} follow discrete-time updates. Therefore, local stability is assessed using the spectral radius of the state-dependent linear term that evolves agent actions in the latent space. The stability analysis module in \methodName{} is applied after training and does not interfere with the training process itself. Thus, it serves as a post-hoc diagnostic tool for pretrained policies, rather than modifying the optimization pipeline.

While formal global stability of linear time-varying systems depends on the joint spectral radius (JSR) of the operator sequence $\{\mathbf{I} + \mathbf{A}_t\}$~\cite{jungers2009joint}, computing the JSR is generally infeasible when $\mathbf{A}_t$ is state-dependent and varies continuously across time. \methodName{} instead performs stability diagnostics locally at fixed state snapshots using the spectral radius of $\{\mathbf{I}+\mathbf{A}_t\}$,  offering a tractable and interpretable view of action evolution in specific regions of state space. Indeed, one may use this formulation to assess the spectral properties of the latent dynamics for various state-action combinations before deploying any pretrained agent. At each time step, the following is computed:
\begin{equation}
    \rho(\mathbf{I}+\mathbf{A}_t) = \max_i \left| \lambda_i(\mathbf{I}+\mathbf{A}_t) \right|,
\end{equation}
Here, \textbf{\emph{local stability}} is guaranteed when the spectral radius of this update matrix is less than one, i.e., $\rho(\mathbf{I}+\mathbf{A}_t) < 1$. In addition to stability, the imaginary components of the eigenvalues of the matrix $\mathbf{A}_t$, denoted $\mathrm{Im}(\lambda_i)$, are analyzed to characterize oscillatory dynamics in the latent space. This helps reveal cycles or damped oscillations in control behavior, which may not be visible through magnitude alone. Specifically, local dynamics are characterized as follows:
\begin{itemize}[nosep]
    \item \(\rho(\mathbf{I}+\mathbf{A}_t) \le 1\) with \(\mathrm{Im}(\lambda_i) = 0\): non-oscillatory locally stable dynamics,
    \item \(\rho(\mathbf{I}+\mathbf{A}_t) \le 1\) with \(\mathrm{Im}(\lambda_i) \neq 0\): locally stable dynamics with damped oscillations,
    \item \(\rho(\mathbf{I}+\mathbf{A}_t) > 1\): locally unstable dynamics, regardless of the imaginary component.
\end{itemize}

Additionally, neighborhood states around the observed states $s_\text{range}$ are evaluated by computing the corresponding $\mathbf{A}_t$ and analyzing their eigenvalues and spectral radius. This reveals locally stable or unstable regions, capturing how state variations influence stability. Importantly, the objective is not to analyze the full closed-loop environment-agent system but to empirically characterize the behavior of the learned policy through its latent action dynamics. This local stability characterization serves as a proxy for inferring how consistent, structured, or potentially unstable the policy's actions are over time, especially in high-stakes or safety-critical applications where local interpretability is crucial.
This also provides insights into regions where the policy achieves stable control or displays unstable behavior. The algorithm is outlined in~\ref{appendix:algorithms}. {\color{black} It is important to note that local stability in the latent action space measures the internal consistency and smoothness of the control policy's decision-making process. It does not imply or guarantee the formal closed-loop stability of the physical system.} 
{\color{black}Empirically, this proxy strongly correlates with actual physical behavior provided the latent representation maintains high fidelity (i.e., minimal reconstruction error) and the environment lacks severe unmodeled disturbances that might decouple control intent from state evolution. Furthermore, the interpretation of near-critical ($\rho \approx 1$) or locally unstable ($\rho > 1$) behavior is highly task-dependent. In static regulation (e.g., hovering), $\rho > 1$ typically indicates a risky regime or imminent control failure, as actions should ideally contract. Conversely, in dynamic balancing (e.g., CartPole), persistent near-critical behavior ($\rho \approx 1$) is an inherent signature of active continuous regulation. Similarly, in limit-cycle locomotion (e.g., walking), periodic transient fluctuations ($\rho > 1$) naturally emerge as nominal, functional dynamics while the system traverses the inherently unstable phases of a sustained locomotion cycle. Finally, it should be noted that this evaluation of physical control regularity is fundamentally orthogonal to epistemic uncertainty estimation (such as value-network ensembles or Bayesian dropout), which quantifies out-of-distribution state familiarity rather than dynamic action consistency.}

\subsection{Transient Growth Analysis}\label{sec:transient}
In the previous section, local stability analysis of latent (i.e., approximate) action dynamics was used as a proxy for assessing the local stability of a trained agent. However, even when the local stability condition $\rho(\mathbf{A}_t) < 1$ is met, non-normality (i.e., $\mathbf{A}_t\mathbf{A}_t^* \neq \mathbf{A}_t^*\mathbf{A}_t$) can still induce transient growth.
This transient amplification, driven by the properties of $\mathbf{A}_t$, can destabilize intermediate states, amplify noise, or disrupt training. Due to the local nature of stability analysis, the transient growth study is essential for ascertaining whether a region in state-action phase space may `escape' into a region of unstable action evolution.

For the latent dynamics described in Equation~\ref{eq:Az}, the spectral radius $\rho(\mathbf{A}_t)$ is first computed to confirm the system stability. Second, for cases where the system is locally stable ($\rho < 1$) but $\mathbf{A}_t$ is non-normal, transient growth is analyzed using the Kreiss constant $\eta(\mathbf{A}_t)$, which measures the potential short-term amplification of the system. The Kreiss constant is defined as:
\begin{equation}
\eta(\mathbf{A}_t) = \sup_{|\zeta| > 1} \frac{|\zeta| - 1}{\|(\mathbf{A}_t - \zeta\mathbf{I})^{-1}\|},
\end{equation}
where $\zeta$ is sampled from the complex plane outside the unit circle (i.e., $|\zeta| > 1$). A high $\eta(\mathbf{A}_t)$ suggests a higher likelihood of transient growth. In such a case, one may observe an agent ultimately performing in an unstable manner even though the starting point of a trajectory has a spectral radius that is less than 1. Algorithm details are provided in~\ref{appendix:algorithms}. Overall, the combination of local spectral radius computation and Kreiss constant analysis provides a qualitative understanding of the locally transient behavior of the system, in the absence of global stability analysis. This ensures an indirect assessment of the robustness of the learned control through the proposed framework.

\subsection{Floquet Analysis for Periodic Stability}\label{sec:floquet}
Floquet analysis \cite{klausmeier2008floquet} is a method specifically designed to evaluate the stability of periodic systems and their behavior over time. It quantifies the evolution of small perturbations in the latent space, capturing their growth, decay, or oscillatory behavior through \textit{Floquet exponents}. This complements the spectral radius and Kreiss constant analyses by focusing on periodic stability, where transient growth alone may not capture recurring instabilities or oscillations. This analysis is applicable when the linear term in the latent dynamical system exhibits periodic trends, as seen in tasks requiring repeated oscillatory actions. An example is a pendulum failing to swing upright and instead oscillating at the bottom, where periodic stability is essential for sustained motion. 

For periodic systems, where the system dynamics repeat at regular intervals such that $\mathbf{A}_{t+dt} = \mathbf{A}_t$, the state transition matrix $\boldsymbol{\Phi}_t$ describes the evolution of perturbations, starting with $\boldsymbol{\Phi}_0 = \mathbf{I}$ (the identity matrix). Its evolution is governed by the dynamic matrix $\mathbf{A}_t$:
\begin{align}
    \dot{\boldsymbol{\Phi}}_t &= \mathbf{A}_t \boldsymbol{\Phi}_t, \\
    \boldsymbol{\Phi}_{t+1} &= \boldsymbol{\Phi}_t + \Delta t \cdot \dot{\boldsymbol{\Phi}}_t.
\end{align}
Here, $\Delta t$ is a time discretization step, set to 1 for simplicity. At the final time horizon $T$, the state transition matrix $\boldsymbol{\Phi}_t$ encodes system dynamics over one period. The eigenvalues of $\boldsymbol{\Phi}_t$, known as the Floquet multipliers $\lambda_i$, are computed, and the corresponding Floquet exponents $\mu_i$ are derived as:
\begin{align}
    \lambda_i &= \text{eigvals}(\boldsymbol{\Phi}_t), \\
    \mu_i &= \frac{\ln(\lambda_i)}{T \cdot \Delta t}.
\end{align}
These exponents provide the following stability insights: 
$\mathrm{Re}(\mu_i) < 0$ indicates exponential decay (stable dynamics), 
$\mathrm{Re}(\mu_i) > 0$ indicates exponential growth (unstable dynamics), and 
$\mathrm{Im}(\mu_i) \neq 0$ implies oscillatory behavior with a frequency proportional to $\mathrm{Im}(\mu_i)$.

In \methodName{} framework, $\mathbf{A}_t$ is directly derived from the state $\mathbf{s}_t$, enabling the period to be identified by analyzing state information. Floquet exponents are computed at the end of each episode to diagnose latent dynamics, revealing instability, oscillations, or unstable growth. This analysis identifies regions of instability, oscillatory behavior, or unstable growth. An intuitive application for this is for periodic systems, such as for controlling the pendulum or cartpole benchmarks in RL. A step-by-step procedure is provided in~\ref{appendix:algorithms}.

{\color{black}Crucially, the step-by-step re-encoding mechanism (Section~\ref{sec:method} enables, rather than invalidates, long-term Floquet analysis. By acting as a predictor-corrector loop, re-encoding corrects for unmodeled physical boundaries, such as environmental action clipping, to prevent open-loop latent drift. This continuous anchoring maintains the system on its true physical limit cycle, allowing the state transition matrix $\boldsymbol{\Phi}_T$ to accurately chain the dynamic operators ($\mathbf{A}_t$) along the realized periodic orbit without being confounded by accumulated numerical errors.}

{\color{black}\subsection{Summary of the \methodName{} Framework.} 
To maximize the practical utility of \methodName{}, we outline a sequential diagnostic workflow for these stability metrics. The spectral radius ($\rho$) serves as the mandatory, continuous diagnostic tool and should be computed at every timestep to assess instantaneous local stability and provide early warnings for erratic behavior. Depending on these primary results, secondary metrics can be deployed for deeper analysis. If the system appears locally stable ($\rho \le 1$) but the dynamic matrix is highly non-normal, the Kreiss constant ($\eta$) serves as a supplementary check to detect vulnerabilities to short-term transient overshoots. Alternatively, for tasks characterized by cyclic or repeating behaviors, such as limit-cycle locomotion, Floquet analysis should be applied over a full trajectory cycle to evaluate the long-term stability of the periodic orbit.

In summary, \methodName{} provides a unified pipeline that bridges continuous DRL optimization with explicit stability diagnostics. By lifting raw actions into a state-dependent latent space governed by the dynamic matrix $\mathbf{A}_t$, the framework preserves standard control performance while unlocking advanced mathematical tools---such as the spectral radius, Kreiss constant, and Floquet multipliers. Together, these techniques systematically evaluate the internal regularity, transient growth, and periodic behavior of the learned policy prior to deployment.}

\section{Experiments}
A range of classical RL environments are selected from OpenAI Gym \cite{brockman2016openai}, focusing on continuous control tasks such as Cartpole-v1 (with continuous action space), Pendulum-v1, LunarLanderContinuous-v2, and BipedalWalker-v3. {\color{black}Complete hyperparameter settings and training steps are detailed in~\ref{appendix:implementation_details}.} Additionally, a modified LunarLander is analyzed with a hovering objective to enhance stability and interpretability (details in~\ref{appendix:lander_reward}).

\subsection{Eigenvalue-based Local Stability Analysis}
\begin{figure}[!ht]
\begin{center}
\centerline{\includegraphics[width=0.8\columnwidth]{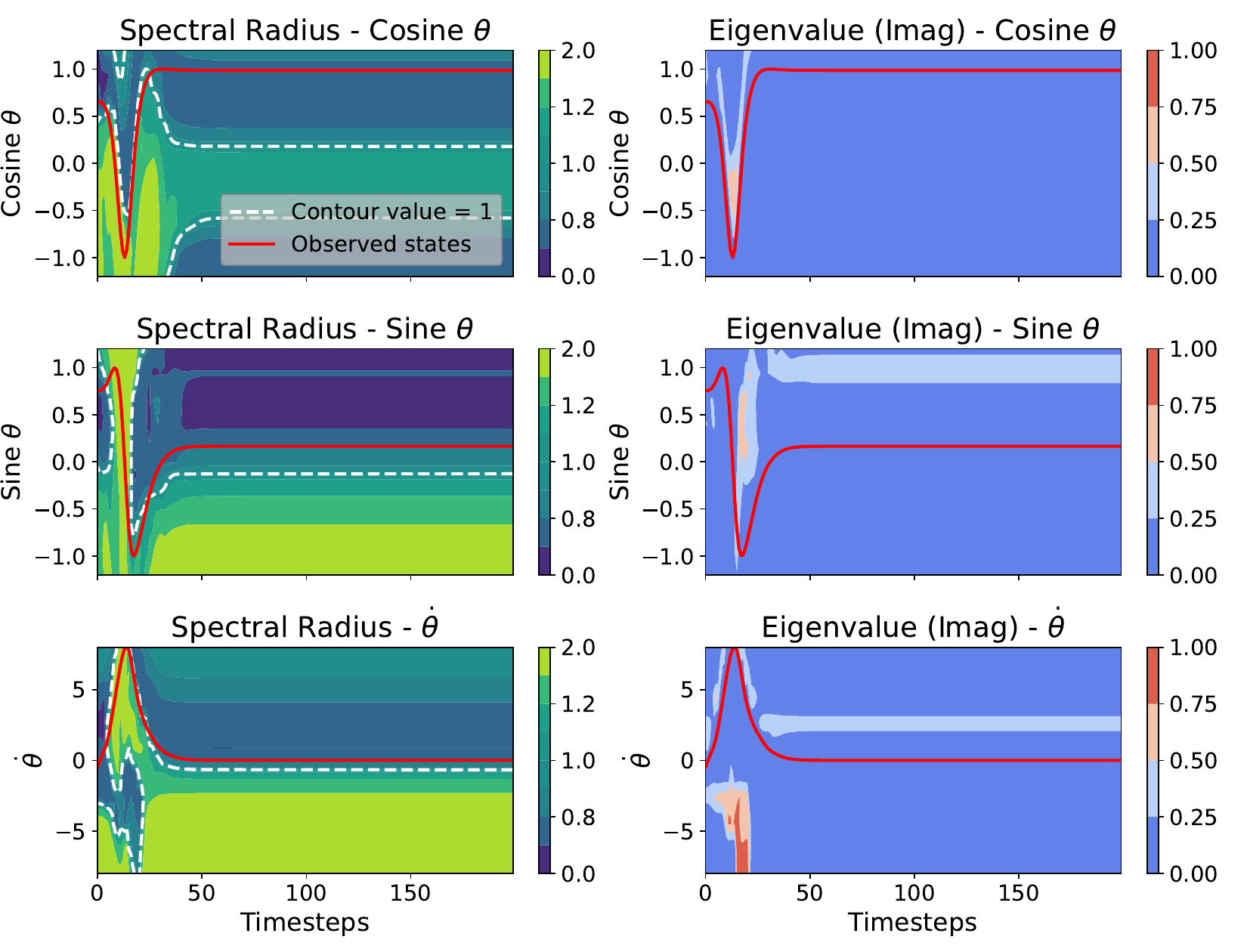}}
\caption{Local stability analysis of Pendulum control. The trajectory (red) transitions from initial regions of growth ($\rho > 1$) into locally contractive zones ($\rho < 1$), maintaining the system within stable bounds defined by the $\rho=1$ boundary.}
\label{fig:pendulum}
\end{center}
\end{figure}

\textbf{Pendulum.} 
This task involves controlling a pendulum using a single torque input at the pivot. The objective is not just to swing the pendulum upright, but to position it in the upright position and maintain balance without further oscillation or swinging. This requires precise and continuous control to counteract gravity. A latent size of $h_d=3$ is used for model training and eigenvalue-based local stability analysis. Figure~\ref{fig:pendulum} shows the system's behavior over time (as a trajectory) and the eigenvalue-based evaluation (underlying contour).

In the first column of the analysis, the trajectory (red line) evolves from locally unstable regions to stable ones. Initially, as the pendulum swings up from the bottom position ($\cos \theta, \sin \theta < 0$), it exhibits instability, reflected by high spectral radius values ($\rho > 1$) in the action space. Over time, the pendulum stabilizes in the upright position ($\cos \theta \to 1$, $\sin \theta \to 0$), with the spectral radius dropping below 1 ($\rho < 1$), indicating entry into a stable regime and steady-state behavior.
In the second column, the imaginary eigenvalues remain at 0, confirming local stability. While states initially approach regions with positive imaginary values, suggesting potential oscillations, they eventually shift away, guiding the system to a steady state. 

In stabilized regimes, it is observed that the latent state continues to evolve, i.e., $\mathbf{z}_{t+1} \neq \mathbf{z}_t$, despite the system being locally stable. This is due in part to the decode–re-encode cycle applied at each step: the predicted latent $\mathbf{z}_{t+1}$ is first decoded into an action $\mathbf{a}_{t+1} = \mathrm{Decoder}(\mathbf{z}_{t+1})$, which is then re-encoded as $\mathbf{z}_{t+1}' = \mathrm{Encoder}(\mathbf{a}_{t+1})$ for the next timestep. The updated latent state then evolves via Equation~\ref{eq:Az} as $\mathbf{z}_{t+2} = (I + \mathbf{A}_t)\mathbf{z}_{t+1}' \approx \mathbf{z}_{t+1}$, causing stabilized decoded actions $\mathbf{a}_{t+2}\approx \mathbf{a}_{t+1}$.

{\color{black}This minor variation occurs because the predicted latent state $\mathbf{z}_{t+1}$ is subjected to physical boundary clipping once decoded into an action. However, because \methodName{} explicitly re-encodes the actually executed action at every time step ($\mathbf{z}_{t+1}' = \mathrm{Encoder}(\mathbf{a}_{t+1})$), any single-step prediction residual is immediately reset. This step-by-step re-encoding mechanism structurally prevents errors from accumulating over time, ensuring that open-loop 'latent drift' cannot occur. Consequently, the decoded actions remain highly stable and effective across time steps, safeguarding long-term control performance.}
This practical invariance in the action space underscores the reliability of the latent local stability diagnostics: even as latent states drift, the resulting control behavior remains consistent, highlighting the model's ability to stabilize behavior through structured latent dynamics.

Additionally, \ref{appendix:results_Pendulum} further examines regions of local instability, recovery, and control failures, reinforcing the framework’s stability analysis. {\color{black}Specifically, this extended analysis qualitatively demonstrates the framework's robustness in detecting extreme distribution shifts and severe unmodeled disturbances, such as intermittent control loss.} Overall, this analysis shows the framework policy not only stabilizes the pendulum but does so in a manner that aligns with the primary objective of keeping it upright. By maintaining eigenvalues within the locally stable region and avoiding oscillatory dynamics, the system ensures stability while providing interpretable insights into how the policy achieves the desired control objectives.  

\begin{figure}[!ht]
\begin{center}
\centerline{\includegraphics[width=0.8\columnwidth]{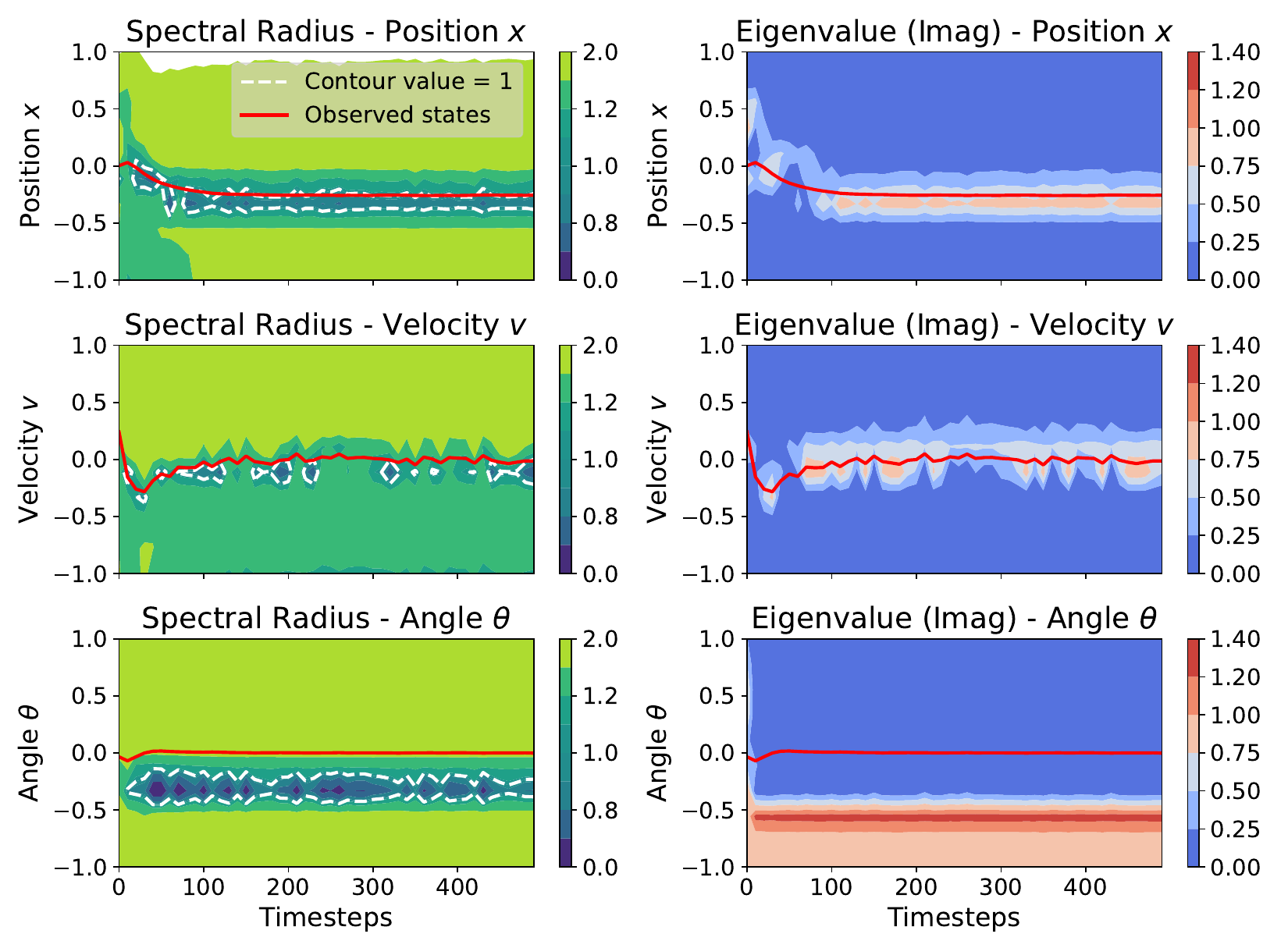}}
\caption{Local stability analysis of CartPole control. Frequent excursions into regions with $\rho(\mathbf{A}_t) > 1$ (red areas) reflect the oscillatory corrective actions required to maintain the pole's upright position. }
\label{fig:cartpole}
\end{center}
\end{figure}

\textbf{CartPole.} The CartPole environment involves a single discrete action that moves the cart left or right to balance the pole. With a latent dimension of $h_d = 3$, the trained model effectively learns to stabilize the pole in the upright position, as shown in Figure~\ref{fig:cartpole}.

However, the spectral radius tends to fluctuate near the white line $\rho=1$, even over extended periods. This behavior can be attributed to the fact that the cart must constantly move left or right to counteract the pole's dynamics. These continual adjustments introduce small, persistent deviations in the latent dynamics, keeping the system near the edge of stability despite overall successful control.
Meanwhile, the imaginary eigenvalue contours reveal a similar pattern of oscillatory behavior. The values consistently fluctuate between zero and nonzero, indicating persistent but bounded oscillations in the latent dynamics throughout the episode.

\begin{figure}[!ht]
\begin{center}
\centerline{\includegraphics[width=1\columnwidth]{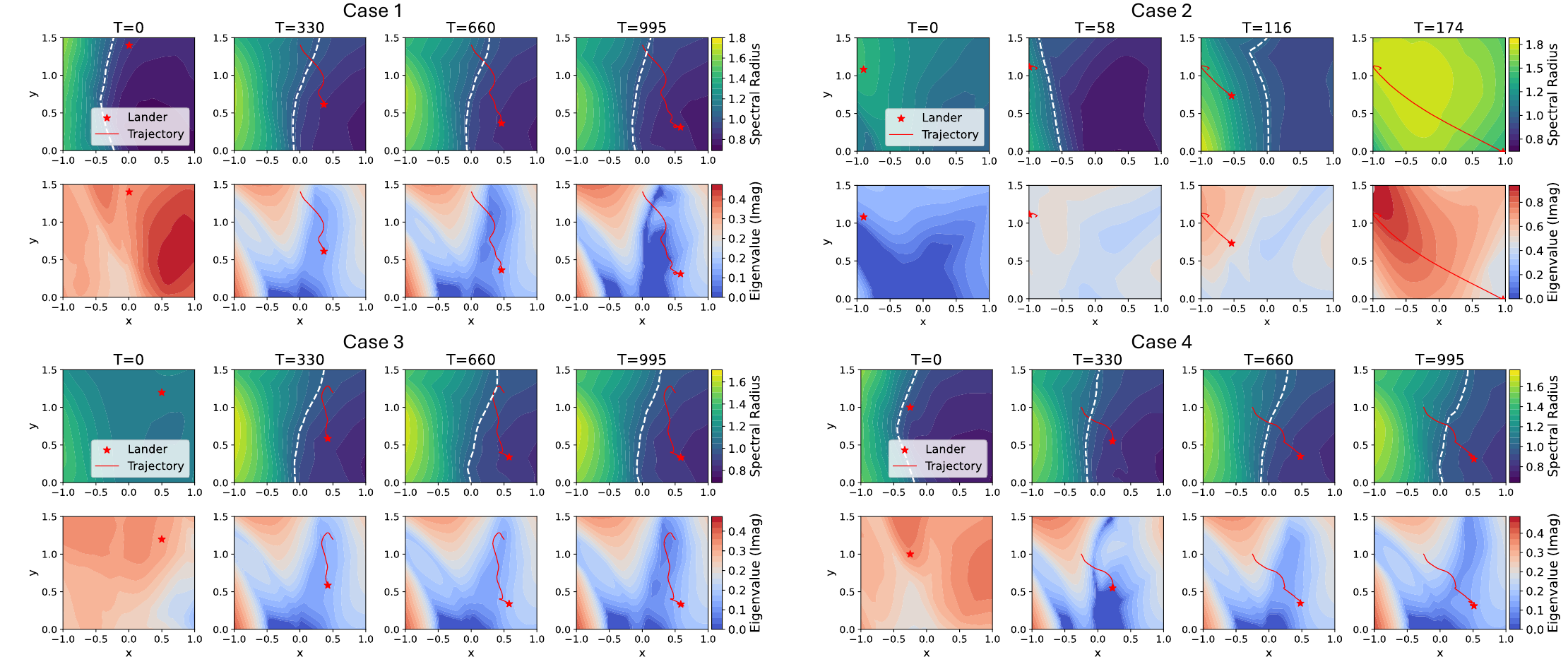}}
\caption{Local stability analysis of LunarLander hovering control across different initializations. $\rho=1$ (white line) delineates locally stable and unstable regions. Trajectories and corresponding eigenvalue distributions demonstrate that the stability contours provide early indicators of potential failure, as observed in the unstable Case 2 before the crash.}
\label{fig:lander}
\end{center}
\end{figure}

\textbf{LunarLander (Hovering Objective).} For the Lunarlander experiment, a modified environment is utilized with a hovering objective instead of the typical objective to land safely. Since the state includes a 2D space, eigenvalue-based local stability is evaluated across the entire 2D domain, with each frame in Figure~\ref{fig:lander} representing a single timestep. 

Among Cases 1, 3, and 4, the spectral radius consistently transitions from high-value regions to lower-value regions, reflecting a transition from locally unstable dynamics to more stable behavior over time. This gradual shift is visually indicated by the trajectory moving from red/yellow zones to cooler regions in the contour plots, especially in the later timesteps. The imaginary contour highlights regions of oscillatory behavior, and the lander's actual trajectory exhibits a transition from high to low oscillation over time.
In case 2, however, the lander crashes at the final frame. The spectral radius contour in this case reveals large regions of local instability, aligning with the lander’s inability to maintain its hovering status. 
Notably, the contour shows that the current state-action combination enters a highly unstable region, effectively signaling the risk of a crash before it occurs.

{\color{black} In \methodName{}, the latent action is not clipped; the clamp is applied only after decoding. Detecting latent-space instability, therefore, pinpoints where the policy relies on saturation rather than manual clipping, as unbounded latent growth forces the decoder to flip between clipped extremes. In this modified LunarLander task, spikes in latent-action instability emerged several steps before any visible state deviation and reliably predicted impending crashes, even though the thrust values themselves remained within their clipped limits.

Overall, this strongly underscores the value of the proposed local stability analysis, as it provides early warnings of imminent failures. Because action stability ultimately drives state evolution, latent action-space analysis yields policy-level interpretability, showing how the agent reacts over time rather than merely its eventual state. Empirically, in the modified LunarLander task, spikes in latent-action instability consistently preceded crashes even while the visible state trajectory still looked nominal. Overall, SALSA-RL is not a replacement for state-based analysis but a complementary, policy-aware diagnostic tool.} {\color{black}To further validate this diagnostic capability, an empirical comparison further demonstrating the robustness of the spectral radius over the highly volatile raw action difference norm is detailed in~\ref{appendix:diff_norm}.}

Besides the above 3 experiments, \ref{appendix:results_walker} explores the BipedalWalker benchmark with additional results and discussions, reinforcing the framework and analysis. {\color{black} Furthermore, \ref{appendix:results_humanoid} extends \methodName{} to the high-dimensional Humanoid environment (featuring a 108-dimensional state and 21-dimensional action space). By appropriately scaling the latent dimension ($h_d \ge 32$) to overcome the information bottleneck observed in lower-dimensional setups, \methodName{} achieves performance comparable to PPO while additionally enabling post-hoc stability analysis. These results conclusively demonstrate the scalability and robustness of the framework, highlighting how the latent-space structure supports local stability analysis and offers increasing benefits in higher-dimensional or more complex control tasks.}


\subsection{Transient Growth Analysis}  
\begin{figure}[ht]
\begin{center}
\centerline{\includegraphics[width=0.8\columnwidth]{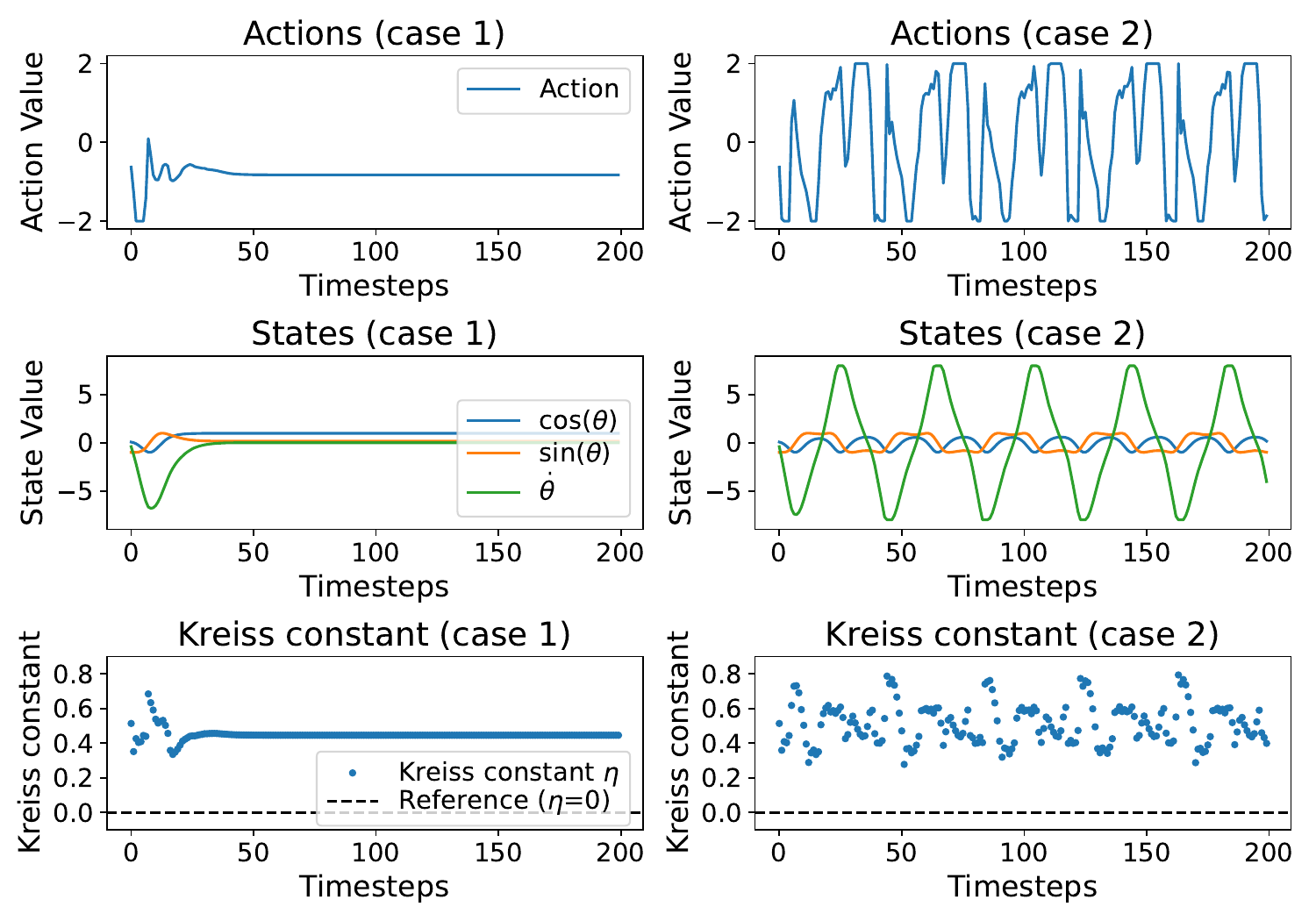}}
\caption{Transient growth and Floquet analysis of pendulum control. Two scenarios are compared: Case 1 (stabilization) and Case 2 (periodic oscillations). The Kreiss constant (bottom row) identifies distinct transient growth characteristics, with corresponding Floquet exponents $\mu_i$ indicating the stability of each regime.}
\label{fig:pendulum_floquet}
\end{center}
\end{figure}

Extending the local stability analysis, the Kreiss constant $\eta$ is evaluated to assess transient growth in the pendulum under a standard setup (Case 1) and a modified environment (Case 2), where gravity increases from 10 to 15 and mass from 1.0 to 1.1. The third row of Figure~\ref{fig:pendulum_floquet} illustrates $\eta$ over time for both cases.  

In Case 1, the pendulum successfully stabilizes in the upright position. This is reflected in the actions and states, which converge to near-zero values without oscillations (first and second rows). The Kreiss constant remains small throughout, indicating minimal transient growth and robust stability. 

In Case 2, the increased gravity and mass disrupt the stabilization process, causing the pendulum to exhibit periodic motion instead of reaching the upright position. This behavior is evident in the actions and states, which display sustained oscillations over time. The Kreiss constant shows periodic spikes, reflecting transient growth that aligns with the observed oscillatory response. These spikes indicate significant susceptibility to transient amplification, even though the system remains bounded over time.

This highlights the role of the Kreiss constant in capturing transient growth, particularly under non-normal dynamics. While spectral radius confirms asymptotic stability, the Kreiss constant reveals critical differences in transient behavior, providing insights into the system’s robustness and susceptibility to perturbations.

\subsection{Floquet Analysis for Periodic Stability}
Building on the two pendulum control cases in Figure~\ref{fig:pendulum_floquet}, Floquet analysis is performed to evaluate periodic stability, providing complementary insights into the stability and oscillatory behavior of the latent dynamics where metrics like eigenvalue analysis or the Kreiss constant may fall short.

In Case 1 (left column of Figure~\ref{fig:pendulum_floquet}), the Floquet analysis reveals all exponents $\mu_i = 0$ across the three-dimensional latent space, indicating marginal stability. This means that small perturbations neither grow nor decay over the period, and the system maintains its oscillatory trajectory without exponential divergence or damping. Zero imaginary components further confirms that the observed oscillations are structural or externally driven, rather than intrinsic to the latent dynamics. 

In Case 2 (right column of Figure~\ref{fig:pendulum_floquet}, modified environment), the $\mu_i=\bigl\{0.96,\ 0.02 \pm 0.04i\bigr\}$ reveal instability, with positive real parts driving perturbation growth and imaginary components introducing oscillatory modes. These dynamics correspond to the pendulum’s observed periodic motion, characterized by high-frequency control oscillations, significant angular velocity peaks, and sustained oscillatory state trajectories.

This analysis demonstrates the utility of Floquet exponents in diagnosing periodic instability and oscillatory behavior, offering deeper insight into latent action dynamics. Moreover, it potentially identifies latent space regions where policy refinement may enhance control and mitigate instability.

\subsection{Ablation Study and Benchmark}
While the goal of this work is not to produce a superior RL algorithm for specific metrics, an ablation analysis is carried out by varying the hidden dimension sizes $h_d$. The results, detailed in Table~\ref{tab:performance} \cite{landajuela2021discovering, rl-zoo3}, demonstrate that this approach achieves interpretability while maintaining performance comparable to state-of-the-art DRL methods and symbolic approaches (DSP) \cite{landajuela2021discovering}. The standard LunarLanderContinuous-v2 is used to ensure consistency in comparisons.
{\color{black}To ensure rigorous evaluation, all SALSA-RL models are evaluated over 1,000 random seeds, with the standard deviation reported alongside the mean rewards.} Only with a severely restricted dimension ($h_d=3$) does the method struggle with high-complexity BipedalWalker, revealing the limitations of a very low-dimensional latent space for effectively encoding the agent's action dynamics. 
Additional ablation studies on {\color{black}computational overhead and stability analysis are detailed in~\ref{appendix:computation_overhead} and ~\ref{appendix:ablation_hd}, respectively}.

{\color{black}Based on these empirical findings, a practical rule-of-thumb for selecting the latent dimension $h_d$ is to always ensure $h_d > \dim(\mathbf{a})$ to prevent information bottlenecking. For standard continuous control tasks, scaling $h_d$ to $2\times$ the action dimension typically provides sufficient degrees of freedom for the dynamic matrix $\mathbf{A}_t$ to capture cross-dimensional couplings. For highly complex, high-dimensional locomotion tasks (e.g., Humanoid), larger dimensions (e.g., $32$ or $128$) are recommended to maintain autoencoder reconstruction fidelity, with the upper bound constrained primarily by computational overhead and the risk of overfitting.}

\begin{table*}[!ht]
\vskip -0.05in
\centering
\renewcommand{\arraystretch}{1.45}
\caption{Average episodic rewards of \methodName{} across varying latent dimensions $h_d$ compared with RL baselines. Results are averaged over 1,000 random seeds using standard benchmarks. Standard LunarLanderContinuous-v2 is used for consistent comparison.}
\label{tab:performance}
\vskip 0.03in
\resizebox{\textwidth}{!}
{
\setlength{\tabcolsep}{3pt}
\begin{tabular}{l ccccc cccccc}
    \toprule
    \multirow{2}{*}{\textbf{Environment}} & \multicolumn{5}{c}{\textbf{\methodName{}}} & \multirow{2}{*}{\textbf{A2C}} & \multirow{2}{*}{\textbf{SAC}} & \multirow{2}{*}{\textbf{DDPG}} & \multirow{2}{*}{\textbf{TD3}} & \multirow{2}{*}{\textbf{PPO}} & \multirow{2}{*}{\textbf{DSP}}\\
    \cmidrule(lr){2-6} 
    & $h_d=3$ & $h_d=4$ & $h_d=6$ & $h_d=8$ & $h_d=16$ & & & & & \\
    \midrule
    Pendulum        & $-149.2 \pm 87$ & $-149.8 \pm 88$ & $-155.8 \pm 87$ & $-157.1 \pm 98$ & $-149.9 \pm 90$ & -162.2  & -159.3 & -169.0  & -147.1 & -154.8 & -160.5\\
    CartPole        & $1000.0 \pm 0$ & $1000.0 \pm 0$ & $1000.0 \pm 0$ & $1000.0 \pm 0$ & $1000.0 \pm 0$ & 1000.0 & 971.8 & 1000.0 & 998.0 & 993.9 & 999.6\\
    LunarLander     & $257.8 \pm 55$ & $268.3 \pm 36$ & $246.1 \pm 59$ & $260.8 \pm 29$ & $242.6 \pm 68$ & 227.1  & 272.7 & 246.2  & 225.4 & 225.1 & 251.7\\
    BipedalWalker   & -      & $235.1 \pm 92$ & $280.4 \pm 40$ & $280.9 \pm 93$ & $262.6 \pm 17$ & 241.0  & 307.3 & 94.2   & 310.2 & 286.2 & 264.4\\
    \bottomrule
\end{tabular}
}
\end{table*}

\section{Conclusion}
This paper introduces \methodName{}, an RL framework for local stability analysis in latent action space, leveraging action dynamics and deep learning. The utility of \methodName{} is demonstrated across various environments, showcasing how control policies achieve stability, avoiding unstable regions, and maintaining stable states. Analysis of unstable regions reveals how key state-action combinations influence system stability, offering insights for improved control actions and reward function designs. Additional numerical analyses of periodic stability and transient growth provide a deeper understanding of underlying control outcomes, further interpreting the relationship between system states and stability.

Moreover, \methodName{}'s latent linear dynamics yield coherent action regions in phase space, reflecting consistent actions in state space (see~\ref{appendix:results_structured_action}). {\color{black} Importantly, the stability analysis module functions as a post-hoc diagnostic: it is applied after training without interfering with optimization, making the approach algorithm-agnostic in principle and broadly applicable across DRL methods. \methodName{}} may also complement safe and stability-focused RL approaches~\cite{gu2024review, brunke2022safe, zhao2023stable, jin2020stability}, by offering post-hoc analysis tools to interpret policy behavior without modifying the training process.

Finally, the goal of \methodName{} is not to outperform on saturated benchmarks but to introduce an interpretable framework for empirical stability analysis of RL policies, capabilities typically lacking in existing RL algorithms.
It is therefore a powerful yet simple tool for understanding the behavior of pretrained agents from various RL algorithms, offering insights into policy behavior, state transitions, and system stability.
{\color{black} Nonetheless, \methodName{} currently focuses on post-hoc stability analysis and does not enforce safety or correct instability during training. Extending the framework to actively guide policy updates, potentially in conjunction with safe RL methods, remains an important direction for future work.
}

{\color{black}Ultimately, the broader research value of \methodName{} lies in bridging the gap between the empirical success of DRL and the rigorous diagnostic needs of physical control systems. By providing an actionable, post-hoc interpretability framework that evaluates the local consistency of control intent, it facilitates more reliable policy validation. This represents a critical step toward trusting and deploying RL agents in real-world, safety-critical environments where predictive behavior diagnostics are just as important as cumulative reward optimization.}

{\color{black}
\textbf{Scope and Limitations of Latent Stability.} Local stability in the latent action space is distinct from formal closed-loop stability of the physical system. The spectral radius serves as a proxy metric for the regularity and consistency of the policy decision-making process. 
Importantly, the implications of latent instability ($\rho > 1$) are highly task-dependent and do not universally equate to policy failure. While emergent instability often precedes failure in static regulation (e.g., hovering), dynamic locomotion tasks fundamentally rely on continuous corrective actions and local instabilities to maintain global physical stability. In such contexts, local instability represents a necessary mechanism for successful control rather than a defect. To differentiate between functional, transient instability and true policy divergence, tracking the temporal evolution of these metrics provides a highly promising solution. Specifically, quantifying the distance from the stability boundary over consecutive time steps to observe if it grows consistently could serve as a robust indicator of impending catastrophic failure.

A locally stable latent action sequence indicates smooth transitions in control intent but does not provide a formal guarantee for state safety. Physical instability may persist under stable latent dynamics in cases of severe unmodeled disturbances or degenerate policies that output stable but physically inappropriate actions. Additionally, extreme nonlinear distortion during decoding or hard actuation limits can decouple latent stability from state evolution. {\color{black}While latent stability is theoretically linked to physical state bounds through Lipschitz continuity, such formal mappings are intractably complex in DRL environments characterized by non-smooth phenomena like rigid-body collisions.} This analysis is therefore a post-hoc diagnostic tool for identifying erratic behavior and must be interpreted alongside state-space metrics to assess total system safety. Furthermore, while \methodName{} provides qualitative early warnings of erratic behavior (e.g., in modified extreme environments), transitioning these instability indicators into a robust online anomaly detection system, quantified by metrics such as lead-time distributions and false-positive rates, remains a highly promising direction for future research.
}

\section{Acknowledgements}
This research used resources of the Argonne Leadership Computing Facility (ALCF), a U.S. Department of Energy (DOE) Office of Science user facility at Argonne National Laboratory, and is based on research supported by the U.S. DOE Office of Science-Advanced Scientific Computing Research Program, under Contract No. DE-AC02-06CH11357. RM and XL acknowledge funding support from the DOE Office of Science, Fusion Energy Sciences program (DOE-FOA-2905, PM-Dr. Michael Halfmoon), and computational resources from the Penn State Institute for Computational and Data Sciences. 


\begin{thebibliography}{10}
\expandafter\ifx\csname url\endcsname\relax
  \def\url#1{\texttt{#1}}\fi
\expandafter\ifx\csname urlprefix\endcsname\relax\def\urlprefix{URL }\fi
\expandafter\ifx\csname href\endcsname\relax
  \def\href#1#2{#2} \def\path#1{#1}\fi

\bibitem{sutton2018reinforcement}
R.~S. Sutton, Reinforcement learning: An introduction, A Bradford Book (2018).

\bibitem{lillicrap2015continuous}
T.~Lillicrap, Continuous control with deep reinforcement learning, arXiv preprint arXiv:1509.02971 (2015).

\bibitem{yu2023reinforcement}
C.~Yu, X.~Zheng, H.~H. Zhuo, H.~Wan, W.~Luo, Reinforcement learning with knowledge representation and reasoning: A brief survey, arXiv preprint arXiv:2304.12090 (2023).

\bibitem{heuillet2021explainability}
A.~Heuillet, F.~Couthouis, N.~D{\'\i}az-Rodr{\'\i}guez, Explainability in deep reinforcement learning, Knowledge-Based Systems 214 (2021) 106685.

\bibitem{kalashnikov2018scalable}
D.~Kalashnikov, A.~Irpan, P.~Pastor, J.~Ibarz, A.~Herzog, E.~Jang, D.~Quillen, E.~Holly, M.~Kalakrishnan, V.~Vanhoucke, et~al., Scalable deep reinforcement learning for vision-based robotic manipulation, in: Conference on robot learning, PMLR, 2018, pp. 651--673.

\bibitem{brockman2016openai}
G.~Brockman, Openai gym, arXiv preprint arXiv:1606.01540 (2016).

\bibitem{haarnoja2018soft}
T.~Haarnoja, A.~Zhou, P.~Abbeel, S.~Levine, Soft actor-critic: Off-policy maximum entropy deep reinforcement learning with a stochastic actor, in: International conference on machine learning, PMLR, 2018, pp. 1861--1870.

\bibitem{fujimoto2018addressing}
S.~Fujimoto, H.~Hoof, D.~Meger, Addressing function approximation error in actor-critic methods, in: International conference on machine learning, PMLR, 2018, pp. 1587--1596.

\bibitem{aastrom2006advanced}
K.~J. {\AA}str{\"o}m, T.~H{\"a}gglund, Advanced PID control, ISA-The Instrumentation, Systems and Automation Society, 2006.

\bibitem{swarnkar2014adaptive}
P.~Swarnkar, S.~K. Jain, R.~K. Nema, Adaptive control schemes for improving the control system dynamics: a review, IETE Technical Review 31~(1) (2014) 17--33.

\bibitem{skogestad2005multivariable}
S.~Skogestad, I.~Postlethwaite, Multivariable feedback control: analysis and design, john Wiley \& sons, 2005.

\bibitem{glanois2024survey}
C.~Glanois, P.~Weng, M.~Zimmer, D.~Li, T.~Yang, J.~Hao, W.~Liu, A survey on interpretable reinforcement learning, Machine Learning (2024) 1--44.

\bibitem{zabounidis2023concept}
R.~Zabounidis, J.~Campbell, S.~Stepputtis, D.~Hughes, K.~P. Sycara, Concept learning for interpretable multi-agent reinforcement learning, in: Conference on Robot Learning, PMLR, 2023, pp. 1828--1837.

\bibitem{chen2021interpretable}
J.~Chen, S.~E. Li, M.~Tomizuka, Interpretable end-to-end urban autonomous driving with latent deep reinforcement learning, IEEE Transactions on Intelligent Transportation Systems 23~(6) (2021) 5068--5078.

\bibitem{mott2019towards}
A.~Mott, D.~Zoran, M.~Chrzanowski, D.~Wierstra, D.~Jimenez~Rezende, Towards interpretable reinforcement learning using attention augmented agents, Advances in neural information processing systems 32 (2019).

\bibitem{hein2018interpretable}
D.~Hein, S.~Udluft, T.~A. Runkler, Interpretable policies for reinforcement learning by genetic programming, Engineering Applications of Artificial Intelligence 76 (2018) 158--169.

\bibitem{xing2023achieving}
J.~Xing, T.~Nagata, X.~Zou, E.~Neftci, J.~L. Krichmar, Achieving efficient interpretability of reinforcement learning via policy distillation and selective input gradient regularization, Neural Networks 161 (2023) 228--241.

\bibitem{zhao2024learning}
T.~Zhao, G.~Li, T.~Zhao, Y.~Chen, N.~Xie, G.~Niu, M.~Sugiyama, Learning explainable task-relevant state representation for model-free deep reinforcement learning, Neural Networks 180 (2024) 106741.

\bibitem{zhao2025self}
A.~Zhao, et~al., Self-referencing agents for unsupervised reinforcement learning, Neural Networks 188 (2025) 107448.

\bibitem{ding2025learning}
X.~Ding, L.~Wang, P.~Koniusz, Y.~Gao, Learning time in static classifiers, arXiv preprint arXiv:2511.12321 (2025).

\bibitem{lyu2019sdrl}
D.~Lyu, F.~Yang, B.~Liu, S.~Gustafson, Sdrl: interpretable and data-efficient deep reinforcement learning leveraging symbolic planning, in: Proceedings of the AAAI Conference on Artificial Intelligence, Vol.~33, 2019, pp. 2970--2977.

\bibitem{nachum2018data}
O.~Nachum, S.~S. Gu, H.~Lee, S.~Levine, Data-efficient hierarchical reinforcement learning, Advances in neural information processing systems 31 (2018).

\bibitem{pateria2021hierarchical}
S.~Pateria, B.~Subagdja, A.-h. Tan, C.~Quek, Hierarchical reinforcement learning: A comprehensive survey, ACM Computing Surveys (CSUR) 54~(5) (2021) 1--35.

\bibitem{florensa2017stochastic}
C.~Florensa, Y.~Duan, P.~Abbeel, Stochastic neural networks for hierarchical reinforcement learning, arXiv preprint arXiv:1704.03012 (2017).

\bibitem{kenny2023towards}
E.~M. Kenny, M.~Tucker, J.~Shah, Towards interpretable deep reinforcement learning with human-friendly prototypes, in: The Eleventh International Conference on Learning Representations, 2023.

\bibitem{xiong2023interpretable}
L.~Xiong, Y.~Tang, C.~Liu, S.~Mao, K.~Meng, Z.~Dong, F.~Qian, Interpretable deep reinforcement learning for optimizing heterogeneous energy storage systems, IEEE Transactions on Circuits and Systems I: Regular Papers (2023).

\bibitem{yarats2021reinforcement}
D.~Yarats, R.~Fergus, A.~Lazaric, L.~Pinto, Reinforcement learning with prototypical representations, in: International Conference on Machine Learning, PMLR, 2021, pp. 11920--11931.

\bibitem{duan2016benchmarking}
Y.~Duan, X.~Chen, R.~Houthooft, J.~Schulman, P.~Abbeel, Benchmarking deep reinforcement learning for continuous control, in: International conference on machine learning, PMLR, 2016, pp. 1329--1338.

\bibitem{lyu2023task}
X.~Lyu, H.~Hu, S.~Siriya, Y.~Pu, M.~Chen, Task-oriented koopman-based control with contrastive encoder, in: Conference on Robot Learning, PMLR, 2023, pp. 93--105.

\bibitem{yeung2019learning}
E.~Yeung, S.~Kundu, N.~Hodas, Learning deep neural network representations for koopman operators of nonlinear dynamical systems, in: 2019 American Control Conference (ACC), IEEE, 2019, pp. 4832--4839.

\bibitem{watter2015embed}
M.~Watter, J.~Springenberg, J.~Boedecker, M.~Riedmiller, Embed to control: A locally linear latent dynamics model for control from raw images, Advances in neural information processing systems 28 (2015).

\bibitem{weissenbacher2022koopman}
M.~Weissenbacher, S.~Sinha, A.~Garg, K.~Yoshinobu, Koopman q-learning: Offline reinforcement learning via symmetries of dynamics, in: International conference on machine learning, PMLR, 2022, pp. 23645--23667.

\bibitem{rozwood2024koopman}
P.~Rozwood, E.~Mehrez, L.~Paehler, W.~Sun, S.~L. Brunton, Koopman-assisted reinforcement learning, arXiv preprint arXiv:2403.02290 (2024).

\bibitem{yousif2023physics}
M.~Z. Yousif, M.~Zhang, Y.~Yang, H.~Zhou, L.~Yu, H.~Lim, Physics-guided deep reinforcement learning for flow field denoising, arXiv preprint arXiv:2302.09559 (2023).

\bibitem{weiner2024model}
A.~Weiner, J.~Geise, Model-based deep reinforcement learning for accelerated learning from flow simulations, Meccanica (2024) 1--18.

\bibitem{garnier2021review}
P.~Garnier, J.~Viquerat, J.~Rabault, A.~Larcher, A.~Kuhnle, E.~Hachem, A review on deep reinforcement learning for fluid mechanics, Computers \& Fluids 225 (2021) 104973.

\bibitem{wang2024joint}
C.~Wang, Y.~Wang, Y.~Yuan, S.~Peng, G.~Li, P.~Yin, Joint computation offloading and resource allocation for end-edge collaboration in internet of vehicles via multi-agent reinforcement learning, Neural Networks 179 (2024) 106621.

\bibitem{zhang2024unified}
Y.~Zhang, L.~Li, W.~Wei, Y.~Lv, J.~Liang, A unified framework to control estimation error in reinforcement learning, Neural Networks 178 (2024) 106483.

\bibitem{yin2025adaptive}
S.~Yin, Z.~Xiang, Adaptive collision avoidance strategy for usvs in perception-limited environments using dynamic priority guidance, Advanced Engineering Informatics 65 (2025) 103355.

\bibitem{chen2025dynamic}
X.~Chen, S.~Yin, Y.~Li, Z.~Xiang, Dynamic path planning for multi-usv in complex ocean environments with limited perception via proximal policy optimization, Ocean Engineering 326 (2025) 120907.

\bibitem{yin2025real}
S.~Yin, Z.~Xiang, Real-time distributed decision-making for simultaneous target assignment and path planning in multiple unmanned surface vehicles, Expert Systems with Applications 279 (2025) 127457.

\bibitem{delfosse2024interpretable}
Q.~Delfosse, H.~Shindo, D.~Dhami, K.~Kersting, Interpretable and explainable logical policies via neurally guided symbolic abstraction, Advances in Neural Information Processing Systems 36 (2024).

\bibitem{verma2018programmatically}
A.~Verma, V.~Murali, R.~Singh, P.~Kohli, S.~Chaudhuri, Programmatically interpretable reinforcement learning, in: International Conference on Machine Learning, PMLR, 2018, pp. 5045--5054.

\bibitem{jiang2019neural}
Z.~Jiang, S.~Luo, Neural logic reinforcement learning, in: International conference on machine learning, PMLR, 2019, pp. 3110--3119.

\bibitem{jin2022creativity}
M.~Jin, Z.~Ma, K.~Jin, H.~H. Zhuo, C.~Chen, C.~Yu, Creativity of ai: Automatic symbolic option discovery for facilitating deep reinforcement learning, in: Proceedings of the AAAI Conference on Artificial Intelligence, Vol.~36, 2022, pp. 7042--7050.

\bibitem{khaled2022framework}
M.~Khaled, K.~Zhang, M.~Zamani, A framework for output-feedback symbolic control, IEEE Transactions on Automatic Control 68~(9) (2022) 5600--5607.

\bibitem{reissig2018symbolic}
G.~Reissig, M.~Rungger, Symbolic optimal control, IEEE Transactions on Automatic Control 64~(6) (2018) 2224--2239.

\bibitem{landajuela2021discovering}
M.~Landajuela, B.~K. Petersen, S.~Kim, C.~P. Santiago, R.~Glatt, N.~Mundhenk, J.~F. Pettit, D.~Faissol, Discovering symbolic policies with deep reinforcement learning, in: International Conference on Machine Learning, PMLR, 2021, pp. 5979--5989.

\bibitem{zolman2024sindy}
N.~Zolman, U.~Fasel, J.~N. Kutz, S.~L. Brunton, Sindy-rl: Interpretable and efficient model-based reinforcement learning, arXiv preprint arXiv:2403.09110 (2024).

\bibitem{liu2021physics}
X.-Y. Liu, J.-X. Wang, Physics-informed dyna-style model-based deep reinforcement learning for dynamic control, Proceedings of the Royal Society A 477~(2255) (2021) 20210618.

\bibitem{banerjee2023survey}
C.~Banerjee, K.~Nguyen, C.~Fookes, M.~Raissi, A survey on physics informed reinforcement learning: Review and open problems, arXiv preprint arXiv:2309.01909 (2023).

\bibitem{alam2021physics}
M.~F. Alam, M.~Shtein, K.~Barton, D.~J. Hoelzle, A physics-guided reinforcement learning framework for an autonomous manufacturing system with expensive data, in: 2021 American Control Conference (ACC), IEEE, 2021, pp. 484--490.

\bibitem{wang2022toward}
R.~Wang, X.~Zhang, X.~Zhou, Y.~Wen, R.~Tan, Toward physics-guided safe deep reinforcement learning for green data center cooling control, in: 2022 ACM/IEEE 13th International Conference on Cyber-Physical Systems (ICCPS), IEEE, 2022, pp. 159--169.

\bibitem{hernandez2023port}
Q.~Hern{\'a}ndez, A.~Bad{\'\i}as, F.~Chinesta, E.~Cueto, Port-metriplectic neural networks: thermodynamics-informed machine learning of complex physical systems, Computational Mechanics 72~(3) (2023) 553--561.

\bibitem{nousiainen2021adaptive}
J.~Nousiainen, C.~Rajani, M.~Kasper, T.~Helin, Adaptive optics control using model-based reinforcement learning, Optics Express 29~(10) (2021) 15327--15344.

\bibitem{jurj2021increasing}
S.~L. Jurj, D.~Grundt, T.~Werner, P.~Borchers, K.~Rothemann, E.~M{\"o}hlmann, Increasing the safety of adaptive cruise control using physics-guided reinforcement learning, Energies 14~(22) (2021) 7572.

\bibitem{cho2019physics}
Y.~Cho, S.~Kim, P.~P. Li, M.~P. Surh, T.~Y.-J. Han, J.~Choo, Physics-guided reinforcement learning for 3d molecular structures, in: Workshop at the 33rd Conference on Neural Information Processing Systems (NeurIPS), 2019.

\bibitem{schulman2017proximal}
J.~Schulman, F.~Wolski, P.~Dhariwal, A.~Radford, O.~Klimov, Proximal policy optimization algorithms, arXiv preprint arXiv:1707.06347 (2017).

\bibitem{jungers2009joint}
R.~Jungers, The joint spectral radius: theory and applications, Vol. 385, Springer Science \& Business Media, 2009.

\bibitem{klausmeier2008floquet}
C.~A. Klausmeier, Floquet theory: a useful tool for understanding nonequilibrium dynamics, Theoretical Ecology 1 (2008) 153--161.

\bibitem{rl-zoo3}
A.~Raffin, Rl baselines3 zoo, \url{https://github.com/DLR-RM/rl-baselines3-zoo} (2020).

\bibitem{gu2024review}
S.~Gu, L.~Yang, Y.~Du, G.~Chen, F.~Walter, J.~Wang, A.~Knoll, A review of safe reinforcement learning: Methods, theories and applications, IEEE Transactions on Pattern Analysis and Machine Intelligence (2024).

\bibitem{brunke2022safe}
L.~Brunke, M.~Greeff, A.~W. Hall, Z.~Yuan, S.~Zhou, J.~Panerati, A.~P. Schoellig, Safe learning in robotics: From learning-based control to safe reinforcement learning, Annual Review of Control, Robotics, and Autonomous Systems 5~(1) (2022) 411--444.

\bibitem{zhao2023stable}
L.~Zhao, K.~Gatsis, A.~Papachristodoulou, Stable and safe reinforcement learning via a barrier-lyapunov actor-critic approach, in: 2023 62nd IEEE Conference on Decision and Control (CDC), IEEE, 2023, pp. 1320--1325.

\bibitem{jin2020stability}
M.~Jin, J.~Lavaei, Stability-certified reinforcement learning: A control-theoretic perspective, IEEE Access 8 (2020) 229086--229100.

\bibitem{kingma2013auto}
D.~P. Kingma, Auto-encoding variational bayes, arXiv preprint arXiv:1312.6114 (2013).

\bibitem{makoviychuk2021isaac}
V.~Makoviychuk, L.~Wawrzyniak, Y.~Guo, M.~Lu, K.~Storey, M.~Macklin, D.~Hoeller, N.~Rudin, A.~Allshire, A.~Handa, G.~State, Isaac gym: High performance gpu-based physics simulation for robot learning (2021).

\end{thebibliography}

\newpage
\appendix
\onecolumn

\section{Encoder-Decoder Training and Discussion}\label{appendix:encoder-decoder}
The encoder and decoder each contain three layers with identical middle layer sizes. The encoder maps the action dimension to the hidden dimension $h_d$, while the decoder performs the reverse operation. Notably, a slightly larger number of neurons was used in the decoder, a common practice to enhance decoding performance and ensure a more effective reconstruction mechanism \cite{kingma2013auto}.

For training, action datasets are generated from trained DLR policies, improving data efficiency compared to uniform sampling, particularly in high-dimensional settings. Training was conducted for 500 epochs with a scheduled learning rate decrease, and the final model achieved a mean squared error (MSE) of approximately $10^{-6}$ to $10^{-7}$ across different $h_d$. 

{\color{black}
To further validate the robustness of the autoencoder and ensure the latent stability analysis remains valid across the entire operational domain, the reconstruction error is comprehensively evaluated. Because the autoencoder operates exclusively on actions, which are strictly bounded by the environment (e.g., $[-1, 1]^m$), the risk of encountering out-of-distribution (OOD) inputs during late-stage training or interference is inherently mitigated. As shown in Figure~\ref{fig:ae_recon_error}, the reconstruction MSE is mapped across a uniform grid of the entire 2D action space for the LunarLander environment. The maximum MSE across all possible valid action combinations remains strictly below $8 \times 10^{-6}$. This uniformly negligible error demonstrates that the latent variable $\mathbf{z}_t$ reliably represents the true action $\mathbf{a}_t$ at all times, thereby guaranteeing the validity of the stability analysis derived from the state-dependent matrix $\mathbf{A}_t$.

\begin{figure}[ht!]
\begin{center}
\includegraphics[width=0.48\columnwidth]{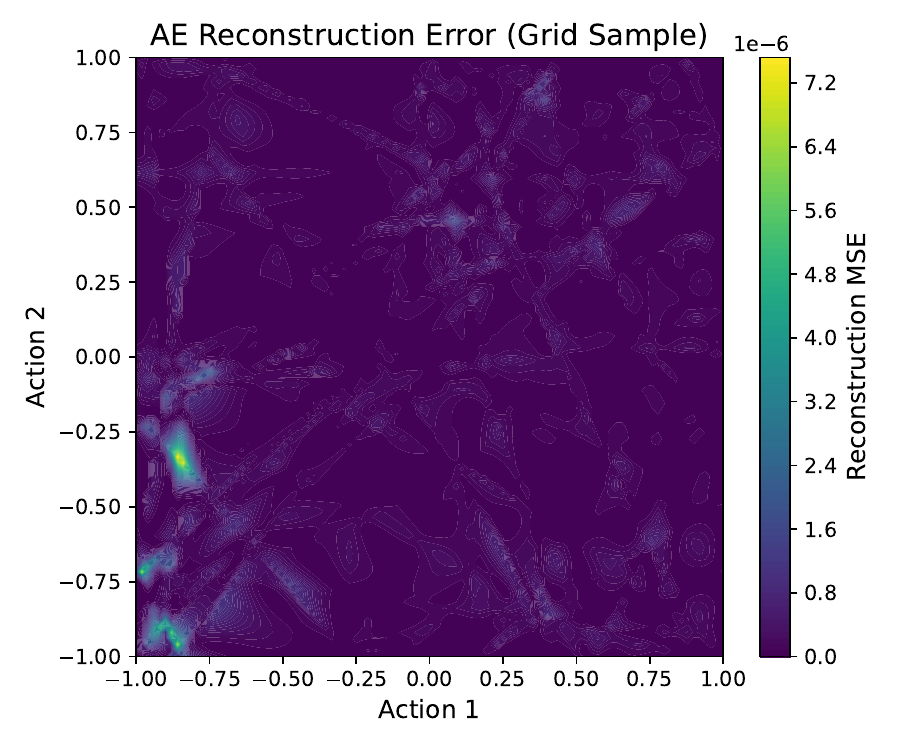}
\caption{Validation of Autoencoder (AE) reconstruction accuracy in the LunarLander environment. A 2D heatmap demonstrating the reconstruction MSE across the entire bounded continuous action space $[-1, 1]^2$. The maximum MSE remains strictly below $8 \times 10^{-6}$ globally, ensuring robust latent stability analysis without out-of-distribution degradation.}
\label{fig:ae_recon_error}
\end{center}
\end{figure}

}
In the BipedalWalker problem, which has four action dimensions, encoding and decoding with a reduced hidden dimension of $h_d=3$ significantly increased the difficulty. Even with a properly scaled network (more neurons and layers), the training MSE loss was three orders of magnitude higher than in other cases. Since precise continuous control is crucial for tasks like BipedalWalker, it is likely that this partially contributed to the failure of \methodName{} to achieve a successful control policy at $h_d=3$ and relatively lower performance at $h_d=4$ (see Table \ref{tab:performance}). However, this should not be viewed as a limitation of \methodName{}, as achieving a compact network size is not the primary objective. In contrast, the stability analysis and interpretability remain consistent across different hidden dimensions, as shown in~\ref{appendix:ablation_hd}.

\section{Derivation of the Policy Gradient for \methodName{}}
\label{appendix:derivation}
\subsection{Policy Formulation}
A latent dynamic control policy is defined based on Eqs.~\eqref{eq:encoder}, \eqref{eq:Az}, \eqref{eq:fs}, and \eqref{eq:decoder}. Substituting the first three equations into the fourth yields a deterministic mapping for generating the next action $\mathbf{a}_{t+1}$  from the current state-action pair $(\mathbf{s}_t, \mathbf{a}_t)$:
\begin{equation}
\label{eq:mu_def}
\begin{split}
\mathbf{a}_{t} &= \pi_\theta(\mathbf{s}_t, \mathbf{a}_{t-1}) \\
      &= \text{Dec}(\mathbf{z}_{t}) \\
      &= \text{Dec}(\mathbf{z}_{t-1}\;+\;\mathbf{A}_t \mathbf{z}_{t-1}) \\
      &= \text{Dec}\bigl(\text{Enc}(\mathbf{a}_{t-1}) \;+\;
        \mathcal{N}_{\theta}(\mathbf{s}_t)\text{Enc}(\mathbf{a}_{t-1})\bigr).
\end{split}
\end{equation}
Here, $\pi_\theta(\cdot)$ is the policy, parameterized by $\theta$, which takes the current state $\mathbf{s}_t$ and previous action $\mathbf{a}_t$ as input. $\mathbf{z}_t$ represents the latent action dynamics at time $t$. $\mathcal{N}$ denotes the latent dynamics network module that takes state $\mathbf{s}_t$ as input and outputs the time-dependent matrix $\mathbf{A}_t$. 

\subsection{Policy Gradient}
Using the deterministic policy gradient (DPG), the gradient for the deterministic policy $\pi_\theta$ is derived, where the objective is to maximize the expected return (cumulative reward), denoted by $J(\pi_\theta)$. The gradient of this objective with respect to the neural network parameters $\theta$ can be written as:
\begin{equation}
\label{eq:dpg_objective}
    \nabla_\theta J(\pi_\theta)
    \;=\;
    \mathbb{E}_{\tau \sim \pi_\theta(\mathbf{a}_t \mid \mathbf{s}_t, \mathbf{a}_{t-1})}
    \Bigl[
        \nabla_\theta \pi_\theta(\mathbf{s}_t, \mathbf{a}_{t-1})
        \;\nabla_a Q^\pi(\mathbf{s}_t,a)\big|_{\,a=\pi_\theta(\mathbf{s}_t,\mathbf{a}_{t-1})}
    \Bigr],
\end{equation}
where $\pi_\theta$ represents the policy parameterized by $\theta$, and $Q$ is the action-value function, which provides the expected cumulative reward starting from state $\mathbf{s}_t$ and taking action $\mathbf{a}_t$.

To apply Eq.~\eqref{eq:dpg_objective} to the proposed latent dynamic control framework, \(\nabla_\theta \pi_\theta(\mathbf{s}_t, \mathbf{a}_{t-1})\) is required. 

Let us define an intermediate latent transformation as:
\begin{equation}
\label{eq:z_forward}
    h_\theta(\mathbf{s}_t, \mathbf{a}_{t-1})
    \;=\;
    \mathbf{z}_{t}
    \;=\;
    \text{Enc}(\mathbf{a}_{t-1})
    \;+\;
    \mathcal{N}_{\theta}\bigl(\mathbf{s}_t\bigr)\text{Enc}\bigl(\mathbf{a}_{t-1}\bigr).
\end{equation}

Then
\begin{equation}
\label{eq:decoder_call}
    \pi_\theta(\mathbf{s}_t,\mathbf{a}_{t-1})
    \;=\;
    \text{Dec}(\mathbf{z}_t)
    \;=\;
    \text{Dec}\bigl(\,h_\theta(\mathbf{s}_t,\mathbf{a}_{t-1})\bigr).
\end{equation}
\noindent \textbf{Chain Rule.} From Eq.~\eqref{eq:dpg_objective}, by the chain rule,
\begin{align}
\label{eq:chain_rule}
    \nabla_\theta \pi_\theta(\mathbf{s}_t,\mathbf{a}_{t-1})
    &=\;
    \frac{\partial\;\text{Dec}(\mathbf{z})}{\partial \mathbf{z}}\Bigg|_{\mathbf{z} = h_\theta(\mathbf{s}_t,\mathbf{a}_{t-1})}
    \;\times\;
    \nabla_\theta \,h_\theta(\mathbf{s}_t,\mathbf{a}_{t-1}) .
\end{align}

\noindent \textbf{Term 1}: $\partial\,\text{Dec}(z)/\partial z$. For Decoder that is fixed/pre-trained, this Jacobian is a constant.
\noindent \textbf{Term 2}: $\nabla_\theta \,h_\theta(\mathbf{s}_t,\mathbf{a}_t)$. From Eq.~\eqref{eq:z_forward},
\begin{align}
    \nabla_\theta \,h_\theta(\mathbf{s}_t,\mathbf{a}_{t-1})
    &=\;
    \nabla_\theta \Bigl[\,
        \text{Enc}(\mathbf{a}_{t-1})
        \;+\;
        \mathcal{N}_{\theta}\bigl(\mathbf{s}_t\bigr)\text{Enc}\bigl(\mathbf{a}_{t-1}\bigr).
    \Bigr]. \notag
\end{align}
Assuming the Encoder is fixed, the only dependence on $\theta$ is through $\mathcal{N}_\theta(\mathbf{s}_t)$. Hence,
\begin{equation}
    \nabla_\theta \,h_\theta(\mathbf{s}_t, \mathbf{a}_{t-1})
    \;=\;
    \text{Enc}(\mathbf{a}_{t-1})
    \;\nabla_\theta \mathcal{N}_\theta(\mathbf{s}_t).
\end{equation}

Putting it all together into Eq.~\eqref{eq:chain_rule}, the following is obtained:
\begin{equation}
    \nabla_\theta \pi_\theta(\mathbf{s}_t, \mathbf{a}_{t-1})
    \;=\;
    \frac{\partial\;\text{Dec}(z)}{\partial z}\Bigg|_{z = h_\theta(\mathbf{s}_t, \mathbf{a}_{t-1})}
    \;\times\; \text{Enc}(\mathbf{a}_{t-1}) \;\nabla_\theta \mathcal{N}_\theta(\mathbf{s}_t).
\end{equation}

Substituting back into Eq.~\eqref{eq:dpg_objective}, the gradient of the objective becomes:
\begin{align}
    \nabla_\theta J(\pi_\theta)
    &= \mathbb{E}_{\tau \sim \pi_\theta(\mathbf{a}_t \mid \mathbf{s}_t, \mathbf{a}_{t-1})}\Biggl[ \notag \\
    &\qquad \Bigl( \frac{\partial\,\text{Dec}(z)}{\partial z}\bigg|_{z=h_\theta(\mathbf{s}_t, \mathbf{a}_{t-1})} \;\times\; \text{Enc}(\mathbf{a}_{t-1}) \;\nabla_\theta \mathcal{N}_\theta(\mathbf{s}_t) \Bigr) \notag \\
    &\qquad \;\nabla_a Q^\pi(\mathbf{s}_t,a)\big|_{a=\pi_\theta(\mathbf{s}_t, \mathbf{a}_{t-1})} \Biggr].
\end{align}
This expression provides the standard DPG update rule for the latent dynamic control framework, where the policy
$\pi_\theta(\mathbf{s}_t, \mathbf{a}_{t-1})$ is derived via the combination of encoder, latent dynamics network $\mathcal{N}_\theta$, and decoder. 

For the stochastic policy, a similar derivation can be performed using the Stochastic Policy Gradient (SPG) theorem, where the gradient involves $\nabla_\theta \log \pi_\theta(\mathbf{a}_t \mid \mathbf{s}_t, \mathbf{a}_{t-1})$ weighted by the action-value function $Q^\pi(\mathbf{s}_t, \mathbf{a}_{t-1})$.

\section{Algorithms for Local Stability Analysis, Transient Growth, and Floquet Analysis} \label{appendix:algorithms}
This section details additional algorithms referenced in the method section of the main article. See details in Algorithm \ref{alg:eigen_evaluation}, \ref{alg:kreiss_analysis}, and \ref{alg:floquet_analysis}.

\begin{algorithm}[!ht]
   \caption{Local Stability Evaluation}
   \label{alg:eigen_evaluation}
\begin{algorithmic}
\STATE {\bfseries Input:} Observed $n$ states $\mathbf{s}_t^{(i)}$, where $i = 1, 2, ..., n$
\FOR{$i = 1$ {\bfseries to} $n$}
    \FOR{$t = 0$ {\bfseries to} $T$}
        \STATE $\hat{\mathbf{s}} \gets [\mathbf{s}_t^{(1)}, \mathbf{s}_t^{(2)}, ..., \mathbf{s}_t^{(n)}]$ // Construct state vector
        \FOR{$s$ {\bfseries in} $s_\text{range}^{(i)}$}
            \STATE $\hat{\mathbf{s}}[i] \gets s$ // Replace with neighboring values
            \STATE $\mathbf{A}_t \gets \mathcal{N}_\theta(\hat{\mathbf{s}})$
            \STATE $\lambda_i \gets \text{eigvals}(\mathbf{A}_t)$ // Eigenvalue calculation
            \STATE $\rho(\mathbf{A}_t) \gets \max_i\left| \lambda_i \right|$ // Spectral radius
            \STATE $\mathrm{Im}_{\max} \gets \max \left( \mathrm{Im}(\lambda_i) \right)$ // Maximum imaginary magnitude
        \ENDFOR
    \ENDFOR
\ENDFOR
\end{algorithmic}
\end{algorithm}

\begin{algorithm}[!ht]
   \caption{Transient Growth Evaluation}
   \label{alg:kreiss_analysis}
\begin{algorithmic}
\STATE {\bfseries Input:} Observed states $\mathbf{s}_t$ over time, error threshold $\epsilon$
\FOR{$t = 0$ {\bfseries to} $T$}
    \STATE $\mathbf{A}_t \gets \mathcal{N}_\theta(\mathbf{s}_t)$
    \STATE $\lambda_1 \gets \text{max}\bigl(\text{eigvals}(\mathbf{A}_t)\bigl)$
    \IF{$\lambda_1 < 1$} 
        \STATE $\text{normality\_diff} \gets \|\mathbf{A}_t \cdot \mathbf{A}_t^\top - \mathbf{A}_t^\top \cdot \mathbf{A}_t\|$
        \IF{$\text{normality\_diff} > \epsilon$} 
            \STATE $\eta(\mathbf{A}_t) \gets \sup_{|z| > 1} \frac{|z| - 1}{\|(\mathbf{A}_t - zI)^{-1}\|}$ // Kreiss constant
        \ENDIF
    \ENDIF
\ENDFOR
\end{algorithmic}
\end{algorithm}

\begin{algorithm}[!ht]
   \caption{Floquet Analysis for Latent Dynamics}
   \label{alg:floquet_analysis}
\begin{algorithmic}
   \STATE {\bfseries Input:} Observed states $\mathbf{s}_t$ over time, state transition matrix $\boldsymbol{\Phi} \gets \mathbf{I}$, identified peak-peak timesteps $t_1$ and $t_2$
   \FOR{$t = 0$ {\bfseries to} $t_2$}
      \STATE $\mathbf{A}_t \gets \mathcal{N}_\theta(\mathbf{s}_t)$
      \IF {$t \ge t_1$}
          \STATE $\boldsymbol{\Phi} \gets \boldsymbol{\Phi} + \mathbf{A}_t \boldsymbol{\Phi}$ // Update state transition matrix
      \ENDIF
   \ENDFOR
   \STATE $\lambda_i \gets \text{eigvals}(\boldsymbol{\Phi}_t)$ // Spectral analysis
   \STATE $\mu_i \gets \frac{\ln(\lambda_i)}{T}$ // Compute Floquet exponents 
\end{algorithmic}
\end{algorithm}

\section{Experimental and Implementation Details}
\label{appendix:implementation_details}
{\color{black}
To ensure full reproducibility, this section provides comprehensive details regarding the network architectures, data collection for the Autoencoder (AE), and the training pipeline used in \methodName{}.

\subsection{Network Architectures}
The detailed configurations of the neural network modules across all evaluated environments are summarized in Table~\ref{tab:network_arch}. All hidden layers utilize the ReLU activation function unless otherwise specified. For the Actor network, which outputs the state-dependent dynamic matrix $\mathbf{A}_t$, a Tanh activation is applied to the final layer, and the output is reshaped to a $h_d \times h_d$ matrix. The Encoder also employs a Tanh activation at its final layer to maintain a bounded latent space representation.

\begin{table}[ht!]
\centering
\caption{Detailed Network Architectures across Environments. The array denotes the number of hidden units in each layer.}
\label{tab:network_arch}
\resizebox{\textwidth}{!}{
\begin{tabular}{l c c c c c}
\toprule
\textbf{Environment} & \textbf{Actor} & \textbf{Critic} & \textbf{AE Encoder} & \textbf{AE Decoder} & \textbf{Latent Dim Tested ($h_d$)} \\
\midrule
Pendulum & [100, 100, 100] & [100, 100, 100] & [40, 40] & [100, 100] & 3, 4, 6, 8, 16 \\
LunarLander & [40, 40, 40] & [40, 40, 40] & [40, 40] & [100, 100] & 3, 4, 6, 8, 16 \\
CartPole & [100, 100] & [100, 100] & [40, 40] & [100, 100] & 3, 4, 6, 8, 16 \\
BipedalWalker & [400, 300] & [400, 300] & [100, 100] & [200, 200] & 4, 6, 8, 16 \\
Humanoid & [400, 200, 100] & [400, 200, 100] & [256, 256] & [256, 256] & 32, 64, 128 \\
\bottomrule
\end{tabular}
}
\end{table}

\subsection{Autoencoder Training Data Source and Scale}
The training data for the action Autoencoder consists of raw action samples collected from the respective environments. For each environment, trajectories are collected spanning 1,500 episodes. The total number of transition steps depends on the maximum episode length of the environment (e.g., approximately 750,000 steps for CartPole and 1,500,000 steps for LunarLander). No additional data scaling or normalization was applied to the action data prior to AE training, as the continuous action spaces defined by the environments are inherently bounded within $[-1, 1]$.

\subsection{Hyperparameter Settings and Training Details}
To ensure fair and rigorous comparisons, the continuous control optimization for \methodName{} was implemented using standard actor-critic architectures across all environments. Both the actor and critic networks were optimized using the Adam optimizer. Depending on the complexity of the environment, the learning rates were tuned ranging from $1 \times 10^{-5}$ to $1 \times 10^{-4}$, with an adaptive learning rate decay mechanism (discount factor of 0.8) utilized in environments requiring finer convergence (e.g., Pendulum). 

Across the primary evaluations, a replay buffer capacity of $1 \times 10^6$ and a batch size of 256 are utilized. The reward discount factor ($\gamma$) was fixed at 0.99. The target networks were updated using soft updates, with the polyak averaging coefficient ($\tau$) selected between 0.001 and 0.005. To encourage sufficient state-space exploration, Gaussian noise was added to the actions during training. The initial standard deviation of the exploration noise was set between 0.2 and 0.4, which was then either decayed over time or clipped (e.g., at 0.5) depending on the strictness of the environment's action bounds. Finally, the total training timesteps were tailored to ensure full convergence, generally spanning from $5 \times 10^5$ to $2 \times 10^6$ steps (e.g., 3,000 episodes for the Pendulum task).

\subsection{Inference Pipeline and Action Clipping}
During both training and inference, \methodName{} follows a strict encode-update-decode sequence as outlined in Algorithm~\ref{alg:1}. Specifically, the latent dynamics update step (i.e., applying the matrix $\mathbf{A}_t$ to $\mathbf{z}_t$) is computed intrinsically without any bounding constraints. Action clipping, which is necessary to match valid environment ranges, is applied strictly to the decoded output prior to environment interaction. Subsequently, this executed action is re-encoded at the next time step to maintain observational consistency. This design ensures that the dynamic matrix $\mathbf{A}_t$ effectively models the local continuous evolution of the action, while the final control output naturally remains compliant with the physical bounds of the environment.

\subsection{\methodName{} Complexity Analysis}
For a standard two-layer actor with state dimension $n$, action dimension $m$, and hidden width $H$, the computational complexity is
\begin{equation}
\mathcal{C}_{\text{Actor}} = O(nH + H^2 + Hm).
\end{equation}

For \methodName{}, the complexity includes the latent dynamics matrix $\mathbf{A}_t$, latent updates, and the encoder/decoder modules:
\begin{equation}
\mathcal{C}_{\text{\methodName{}}} = O(nH + H^2 + H h_d^2 + h_d^2 + 2(mc + c^2 + ch_d)),
\end{equation}
where $h_d$ is the latent dimension and $c$ is the hidden width of the two-layer encoder/decoder.

Since $h_d \ll H \approx c$ in all experiments (e.g., $h_d \leq 16$ while $H, c \approx 200$), the additional terms are negligible relative to the dominant $H^2$ term of the baseline. Hence, the overall complexity of \methodName{} simplifies to
\begin{equation}
\mathcal{C}_{\text{\methodName{}}} \approx O(H^2 + H h_d^2).
\end{equation}
}

\section{Modified LunarLander Environment for Hovering\label{appendix:lander_reward}}
In the standard LunarLanderContinuous-v2 environment, the reward function encourages efficient and accurate landings. The reward is shaped to penalize excessive fuel consumption, provide bonuses for leg contact with the ground, and reward proximity to the flat landing area. Additionally, a large bonus (+100) is given for a successful landing, while a penalty (-100) is applied for crashes or going out of bounds. This design promotes controlled and fuel-efficient landings by balancing penalties and rewards.

In the modified environment, the focus shifts from landing to hovering consistently. The bonuses for leg contacts and successful landing are removed, and a large negative penalty of -300 is applied for crashes, going out of bounds, or touching the ground. A shaping term adjusts the vertical coordinate by 0.2 to encourage hovering slightly above the surface: $\text{shaping} = -100 \sqrt{x^2 + (y - 0.2)^2}.$ A survival bonus of +0.2 per timestep is introduced to reward prolonged hovering, while penalties for fuel consumption remain unchanged to minimize unnecessary actions. These modifications prevent premature convergence to local reward maxima dominated by landing or crashing. Additionally, the larger crash penalty strongly discourages failures, promoting stability and long-term control instead.

\section{Additional Experimental Results}
\subsection{Extended Stability Analysis and Interpretability on Pendulum Problem} \label{appendix:results_Pendulum}
\begin{figure}[ht!]
\begin{center}
\centerline{\includegraphics[width=1\columnwidth]{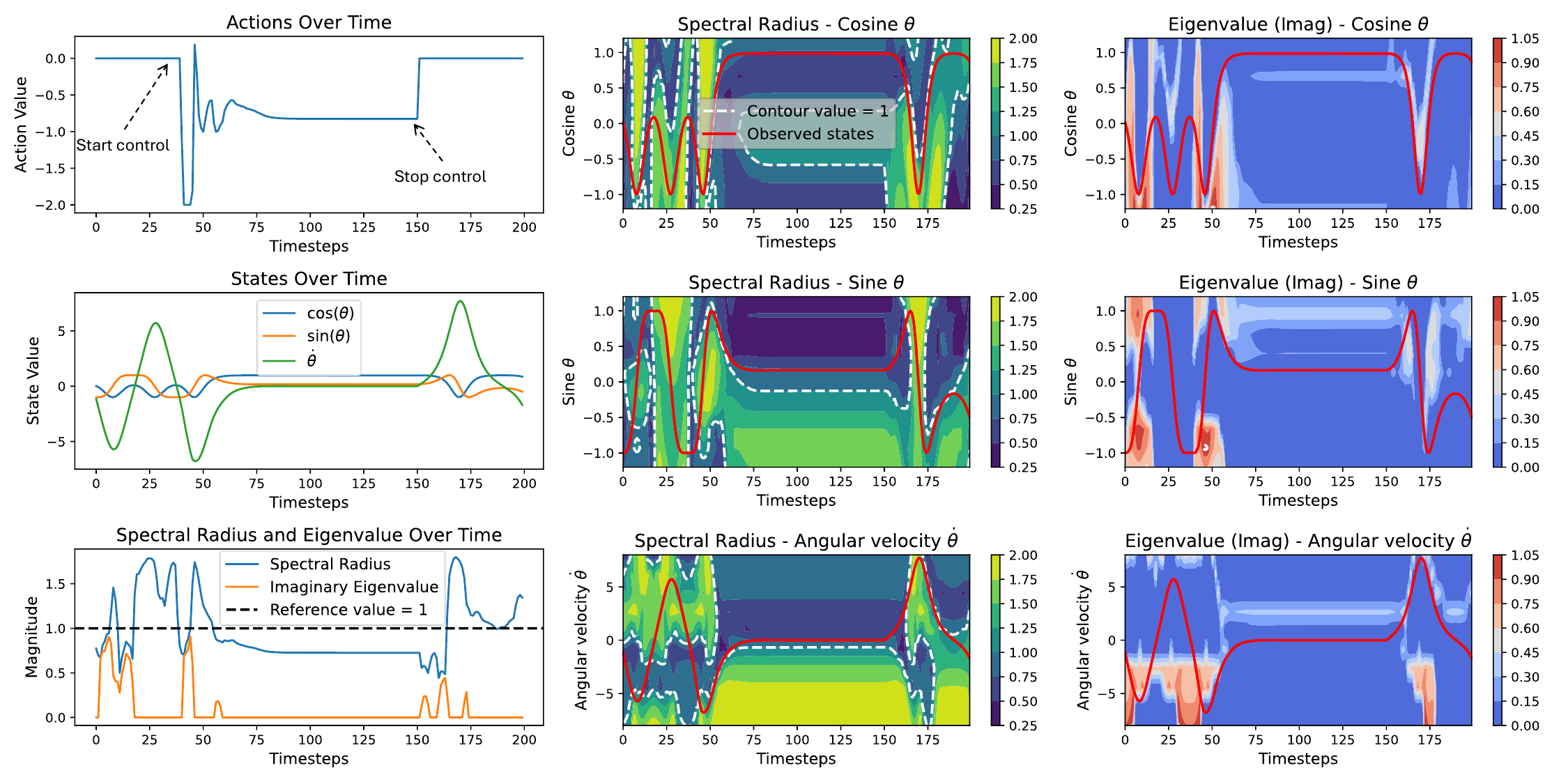}}
\caption{Stability analysis of a custom pendulum scenario with zero-action intervals ($t < 30$ and $t > 150$). The plot shows the evolution of action, state, and stability metrics (spectral radius and imaginary parts), supplemented by phase-space stability contours.}
\label{fig:appendix_pendulum}
\end{center}
\end{figure}

In the main article, the analysis of pendulum control highlighted regions of local stability along with the control actions over time. In Figure \ref{fig:appendix_pendulum}, extreme scenarios involving the absence of control actions are examined as potential control failures, aiming to showcase extended stability analysis during these failure processes.

The environment was randomly initialized, and actions were restricted from timestep 30 to 150. This setup enables the analysis of three typical time regions:
\begin{enumerate}
    \item \textbf{Region of Instability (T: 0--30):} Initial local instability before any actions are applied.
    \item \textbf{Region of Recovery (T: 30--150):} Control actions are applied to recover the system from local instability to stability.
    \item \textbf{Region of Control Failure (T: 150--200):} The control is removed, and the system transitions back from local stability to instability.
\end{enumerate}

From the spectral radius and eigenvalue plot (first column, third row), regions of local instability ($\rho_1>1$) are evident in time Regions 1 and 3, as well as at the initial stage of Region 2. In Region 2, the policy successfully recovers the system from local instability to stability, aligning with the above-defined phases and the pendulum control objective. 
For the imaginary part of the eigenvalue evolution, pronounced oscillatory behavior is observed in Regions 1 and 3, which aligns with the corresponding periods of local instability identified by the spectral radius analysis.

In the second and third columns of the figure, contour plots further validate this observation. Regions of local instability or oscillation near the trajectory are visibly highlighted in yellow and red. These appear on both the left and right sides of the contour plots, corresponding to time regions 1 and 3, and reinforce the established local stability analysis and definition. Overall, this analysis significantly enhances the understanding of control failures, showcasing the interpretability and robustness of the proposed framework.

\subsection{Local Stability Analysis for the BipedalWalker Control}\label{appendix:results_walker}
\begin{figure}[ht!]
\centering
\includegraphics[width=0.49\textwidth]{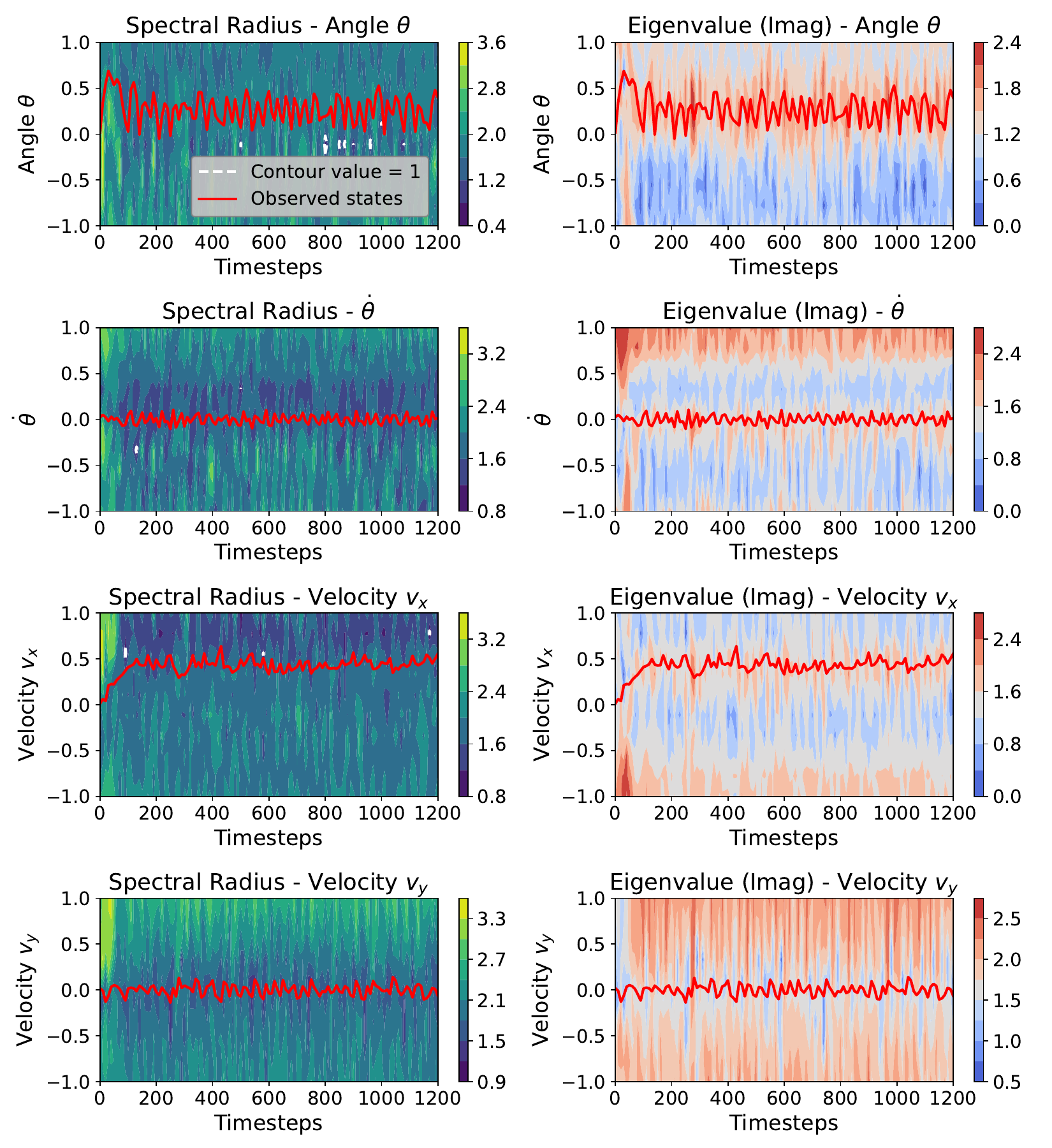}
\hfill
\includegraphics[width=0.49\textwidth]{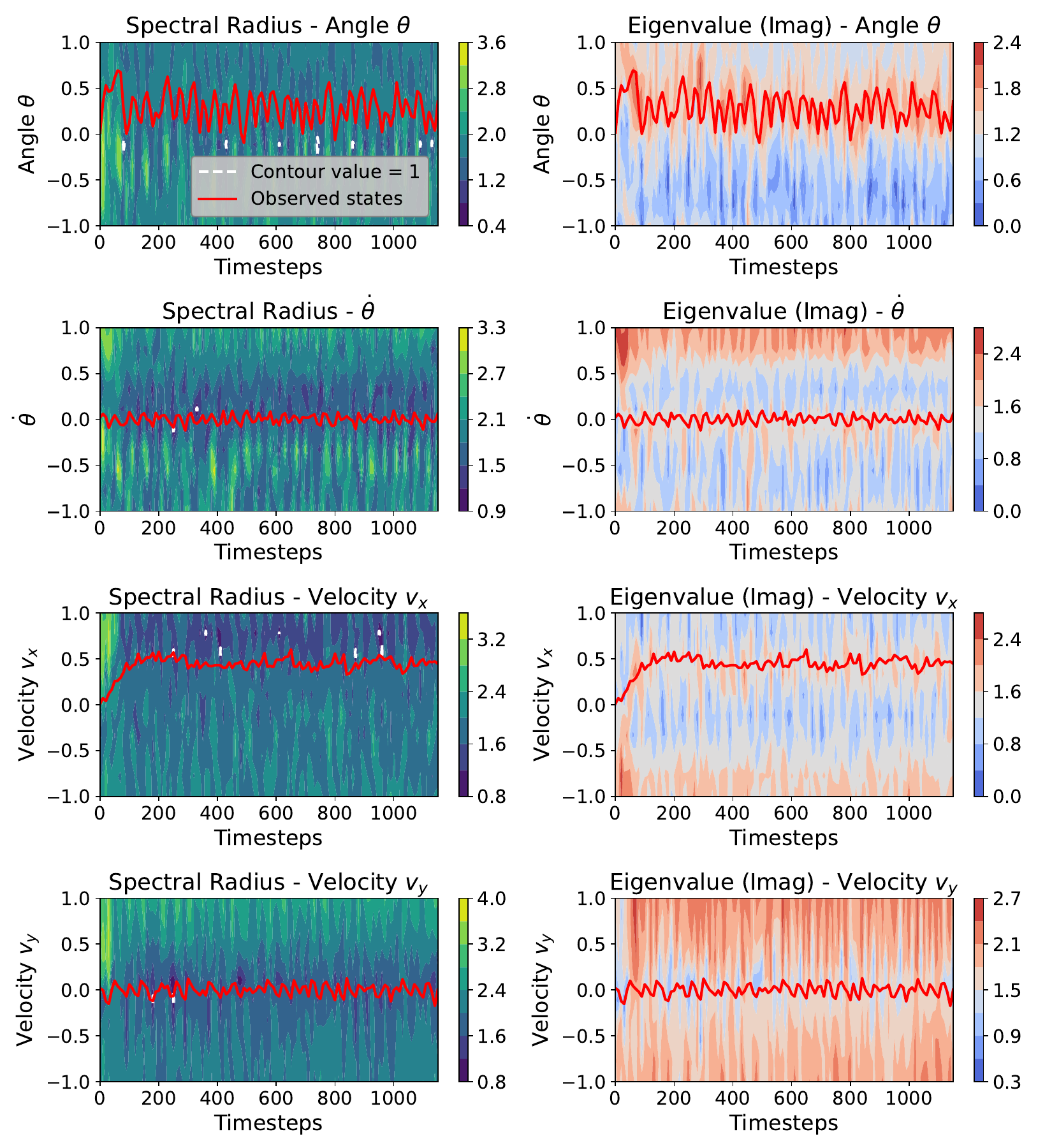}
\caption{Local stability analysis of Bipedal Walker control for two representative cases. Observed trajectories (red lines) show the hull maintaining a positive angle with horizontal progression. High $\rho$ ($>1$) and corresponding imaginary eigenvalue contours indicate that the local instability is primarily driven by periodic oscillations during movement.}\label{fig:walker}
\end{figure}

The Bipedal Walker problem involves numerous states and four actions controlling the legs and joints, aiming to achieve stable, energy-efficient forward locomotion across uneven terrain without falling. Similar to other environments analyzed in the main article, the local stability analysis is presented for two different cases, as shown in Figure \ref{fig:walker}. Due to the large number of states, only four representative states associated with the walker's hull are illustrated.

In both cases, the hull primarily exhibits local instability throughout most of the episode, with the spectral radius remaining significantly greater than 1. These high values are driven largely by the magnitude of the imaginary components of the eigenvalues, even though the real parts remain mostly below 1. From the walker's perspective, this instability, caused by persistent oscillations, corresponds to the agent’s continuous rightward movement. The oscillatory behavior of the hull (i.e., vertical up-and-down motion) emerges naturally as the legs alternate between stance and swing phases. This is also evident in the angle and angular velocity states (shown in the first and second rows), where significant variations in the hull's orientation are observed.

Additionally, during movement, the vertical velocity of the hull (shown in the last row of Figure~\ref{fig:walker}) is minimized and remains close to zero. This behavior is also reflected in the imaginary eigenvalue contour, where the red trajectory resides in a region of low imaginary components (depicted in blue). However, oscillations during movement cannot be entirely eliminated, resulting in non-zero imaginary eigenvalues and, consequently, a system characterized by $\rho > 1$ and continuous use of actions for controlling the system. This example reflects how the spectral radius must be interpreted in the context of the state-action behavior of a system being controlled.

\subsection{Stability Analysis on High-Dimensional Humanoid Environments}
\label{appendix:results_humanoid}

To further assess robustness, \methodName{} is extended to high-dimensional continuous control, specifically the Humanoid task in Isaac Gym~\cite{makoviychuk2021isaac}. This environment features an action space of 21 dimensions and an observation space of 108 dimensions.

For the autoencoder, models are trained with latent dimensions $h_d$ in higher values of $\{32, 64, 128\}$, since lower values significantly degrade reconstruction performance. Both the encoder and decoder are implemented as three-layer neural networks with 256 hidden units per layer.

For reinforcement learning, PPO is adopted with a policy network consisting of three layers of 400, 200, and 100 hidden units, respectively. Training is performed for 3000 epochs. Standard PPO achieves an average reward of approximately 10447 for 1000 timesteps. In comparison, \methodName{} achieves 8999 for $h_d=32$, 9390 for $h_d=64$, and 9170 for $h_d=128$.

\begin{figure}[ht!]
    \centering
    \includegraphics[width=0.85\textwidth]{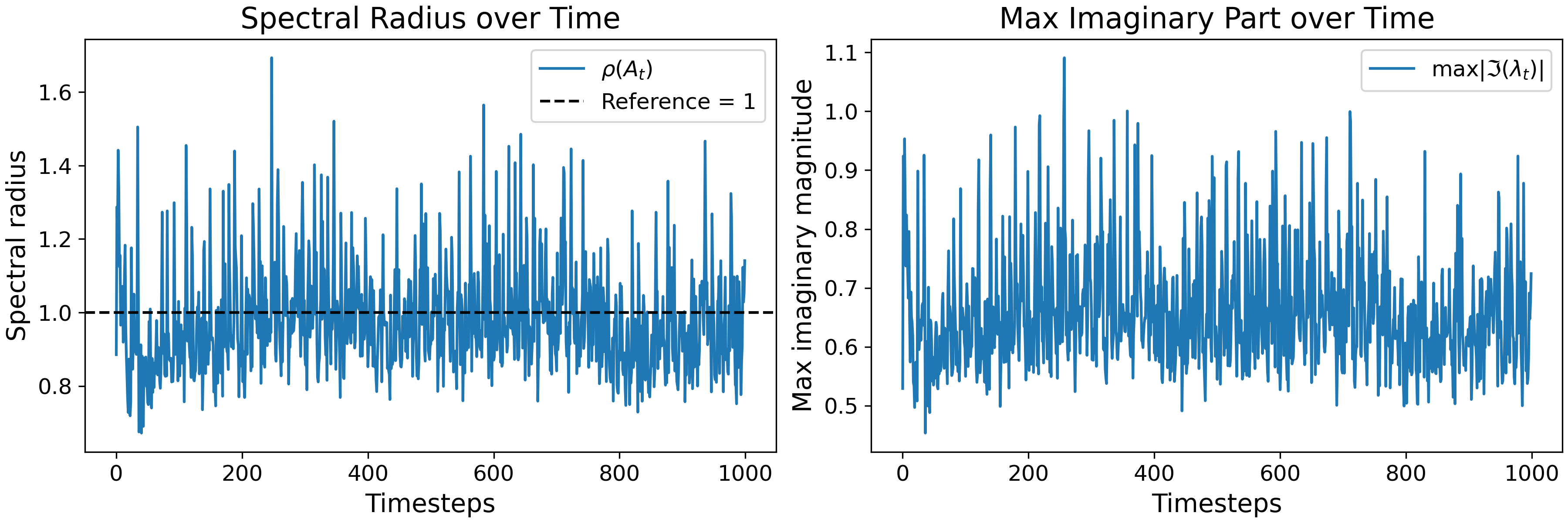}
    \hfill
    \includegraphics[width=0.95\textwidth]{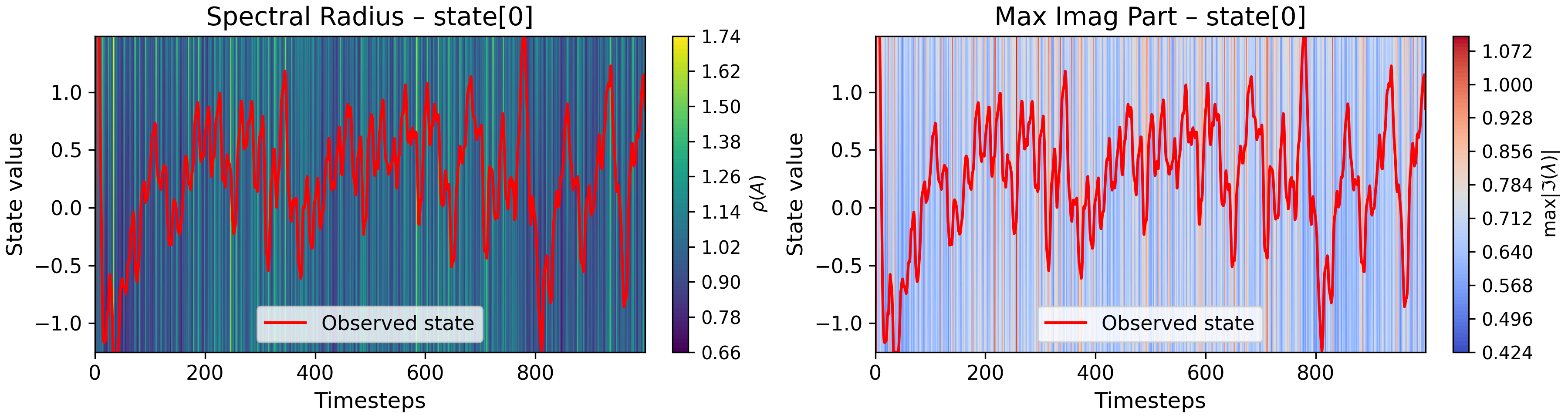}
    \caption{Stability analysis for Humanoid with $h_d=32$. 
    Top row: evolution of actions, states, spectral radius, and imaginary parts of eigenvalues. 
    Bottom row: extended analysis via stability contour visualizations.}
    \label{fig:humanoid}
\end{figure}

Representatively, the spectral radius and imaginary eigenvalue trajectories over time are shown for the case $h_d=32$, as illustrated in Figure~\ref{fig:humanoid}. The spectral radius fluctuates around the unit boundary, with frequent crossings above and below $\rho(\mathbf{A}_t)=1$. This indicates that the learned policy operates at the edge of stability, exhibiting persistent transitions between stable and unstable regimes in the latent action dynamics. While transient excursions into instability are observed, the spectral radius tends to remain close to 1, suggesting that the overall behavior of the policy is near-critical but not dominated by instability.

In the extended analysis, the stability contours over time are visualized for the first dimension of the observation space (e.g., the first state variable), as shown in the second row of Figure~\ref{fig:humanoid}. These plots indicate that, beyond growth rates captured by the spectral radius, imaginary components make notable contributions, reflecting oscillatory modes that influence the observed stability transitions.

Overall, the humanoid results show that \methodName{} achieves comparable performance to PPO while additionally enabling post-hoc stability analysis in high-dimensional settings. This demonstrates the scalability and robustness of the framework, highlighting its potential for analyzing complex control policies beyond low-dimensional benchmark tasks.

{\color{black}
\section{Comparison of Latent Spectral Radius and Raw Action Difference Norm}
\label{appendix:diff_norm}
To explicitly justify the use of the latent spectral radius $\rho(\mathbf{A}_t)$ over simple raw action difference metrics (e.g., $||\mathbf{a}_t - \mathbf{a}_{t-1}||$), a temporal comparison is conducted over a complete trajectory. 

As illustrated in Figure~\ref{fig:diff_norm_vs_rho}, while the two metrics exhibit macroscopic correlation, their measured statistical correlation is notably weak (a correlation coefficient of 0.26).
This extreme noise makes it practically infeasible to establish reliable thresholds for anomaly detection, as minor environmental perturbations trigger sharp spikes resulting in severe false positives. 

Conversely, the spectral radius effectively filters out instantaneous action noise, providing a smooth and coherent signal that robustly captures the underlying stability of the policy's intent. By preserving cross-dimensional couplings through the matrix $\mathbf{A}_t$, the latent dynamic metric enables predictive and granular diagnostics that a simple aggregated scalar difference norm cannot offer.

\begin{figure}[ht!]
\begin{center}
\centerline{\includegraphics[width=0.85\columnwidth]{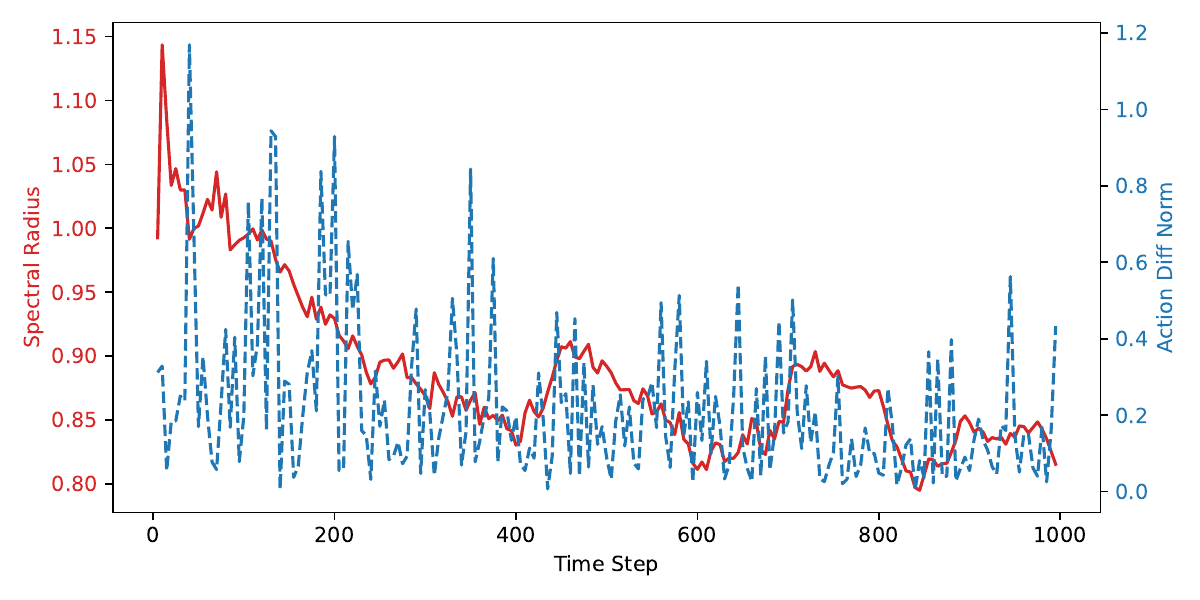}}
\caption{Temporal comparison between the latent spectral radius (red, solid line) and the raw action difference norm (blue, dashed line) over a single trajectory. The spectral radius provides a smooth, coherent indicator of policy stability, whereas the difference norm is heavily dominated by high-frequency noise, making it unsuitable for robust threshold-based diagnostics.}
\label{fig:diff_norm_vs_rho}
\end{center}
\end{figure}
}

\section{Additional Ablation Studies}
{\color{black}
\subsection{Computational Overhead and Inference Latency}\label{appendix:computation_overhead}
A quantitative profiling of inference latency was conducted to evaluate the computational complexity introduced by the latent action dynamics. Measurements were performed in the Pendulum environment using a single NVIDIA A40 GPU, comparing the standard SAC baseline against \methodName{} across varying latent dimensions ($h_d$).

As shown in Table~\ref{tab:inference_time}, results from 10,000 forward passes indicate that the state-dependent matrix $\mathbf{A}_t$ introduces no significant computational bottleneck. The standard SAC baseline (evaluated via RL Baselines3 Zoo \cite{rl-zoo3}) exhibits a latency of $0.4627$ ms, primarily due to software-level abstractions within the library. Conversely, profiling \methodName{} using pure PyTorch forward passes isolates the mathematical overhead of the dynamic matrix.

The inference latency for \methodName{} remains stable at approximately $0.36$ ms per step. Increasing the latent dimension from $h_d = 3$ to $h_d = 16$ results in a maximum latency fluctuation of less than $0.003$ ms, representing a variance under $1\%$. On modern hardware, these matrix operations are highly parallelizable and contribute a negligibly small and effectively constant empirical overhead relative to standard MLP operations, satisfying the requirements for high-frequency real-time control.

GPU memory analysis aligns with these findings (Table~\ref{tab:inference_time}). The baseline SAC model requires approximately $9.17$ MB, while the integration of the latent dynamic matrix in \methodName{} increases this requirement by only $0.2$ MB, totaling $9.37$ MB at $h_d = 16$. This minimal memory footprint demonstrates that \methodName{} imposes no prohibitive resource constraints, facilitating deployment on standard or resource-constrained hardware.
\begin{table}[ht!]
\vskip -0.05in
\centering
\renewcommand{\arraystretch}{1.3}
\caption{Quantitative comparison of per-step inference latency and GPU memory usage for standard SAC and \methodName{} across different latent dimensions ($h_d$) on an NVIDIA A40 GPU.}
\label{tab:inference_time}
\vskip 0.03in
\resizebox{0.95\columnwidth}{!} 
{
\setlength{\tabcolsep}{6pt}
\begin{tabular}{l c c}
\toprule
\textbf{Model Architecture} & \textbf{Inference Latency (ms/step)} & \textbf{GPU Memory (MB)} \\
\midrule
Standard SAC Baseline \cite{rl-zoo3} & 0.4627 & 9.1685 \\ 
\midrule
\textbf{\methodName{}} & & \\
\quad $h_d = 3$ & 0.3569 & 9.2642 \\ 
\quad $h_d = 4$ & 0.3585 & 9.2676 \\
\quad $h_d = 6$ & 0.3578 & 9.2759 \\
\quad $h_d = 8$ & 0.3601 & 9.2876 \\
\quad $h_d = 16$ & 0.3585 & 9.3657 \\
\bottomrule
\end{tabular}
}
\end{table}
}

\subsection{Ablation Study on Stability Analysis Across Different Hidden Dimensions}\label{appendix:ablation_hd}
\begin{figure}[ht!]
    \centering
    \includegraphics[width=0.49\textwidth]{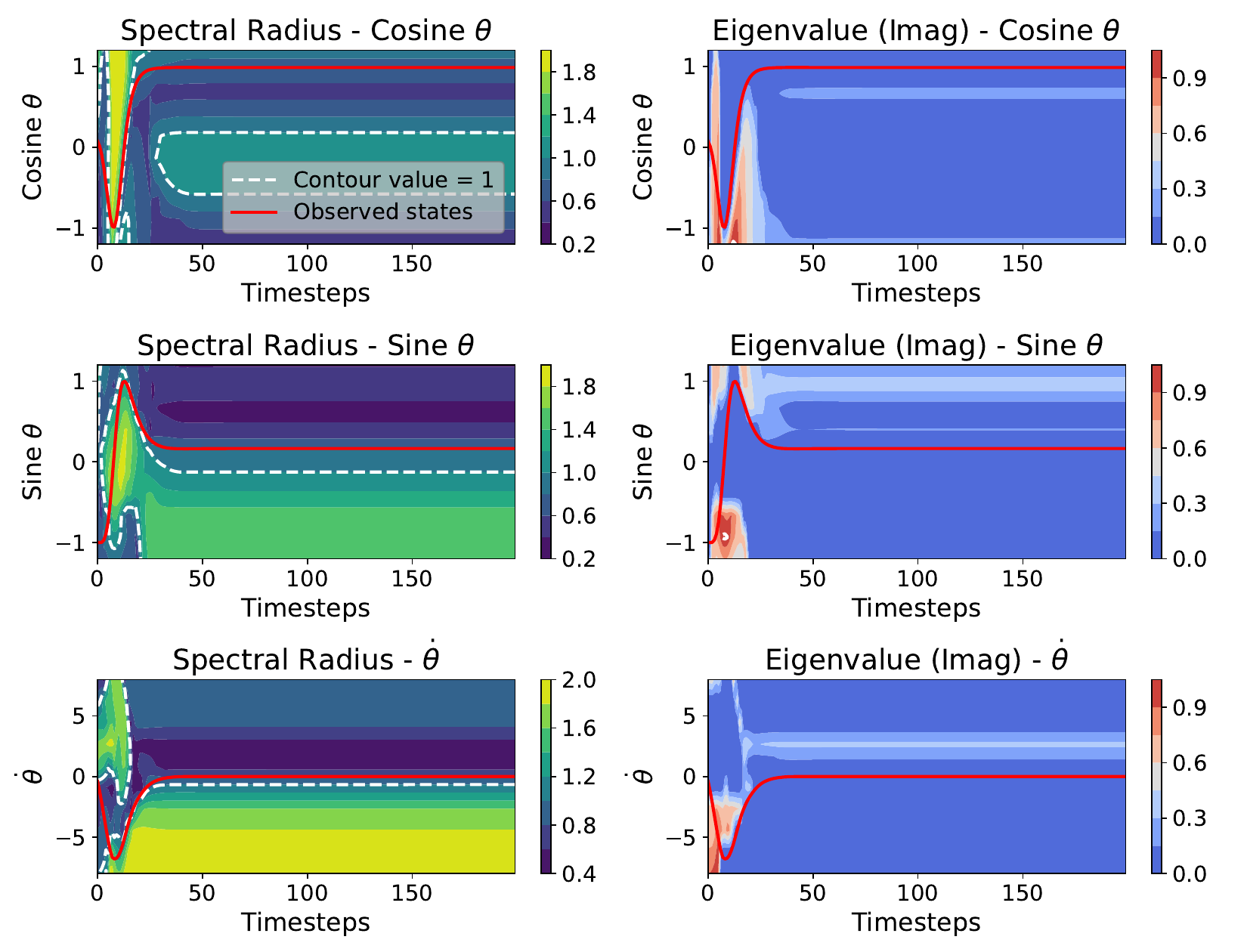}
    \hfill
    \includegraphics[width=0.49\textwidth]{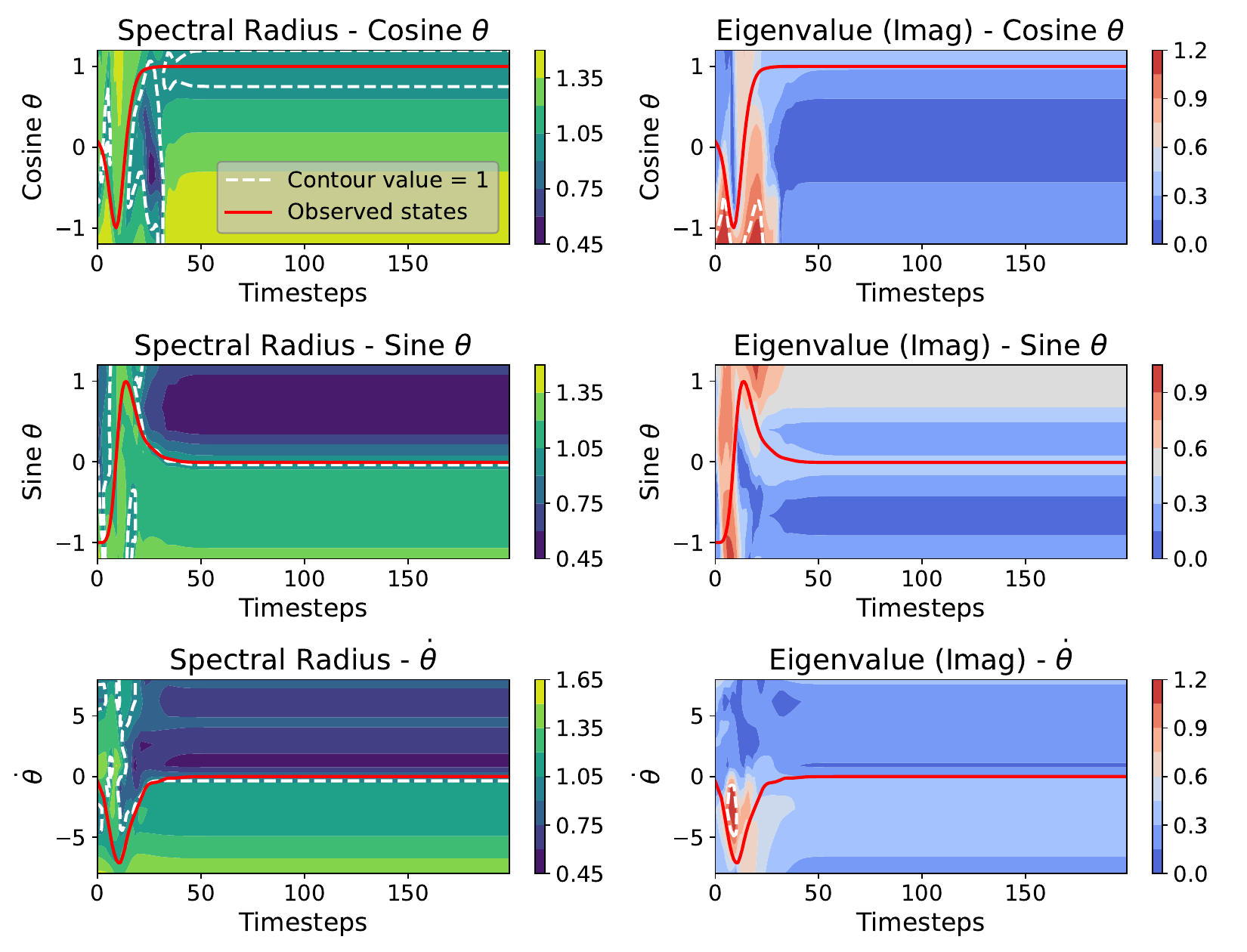}
    \vskip 0.2in
    \includegraphics[width=0.49\textwidth]{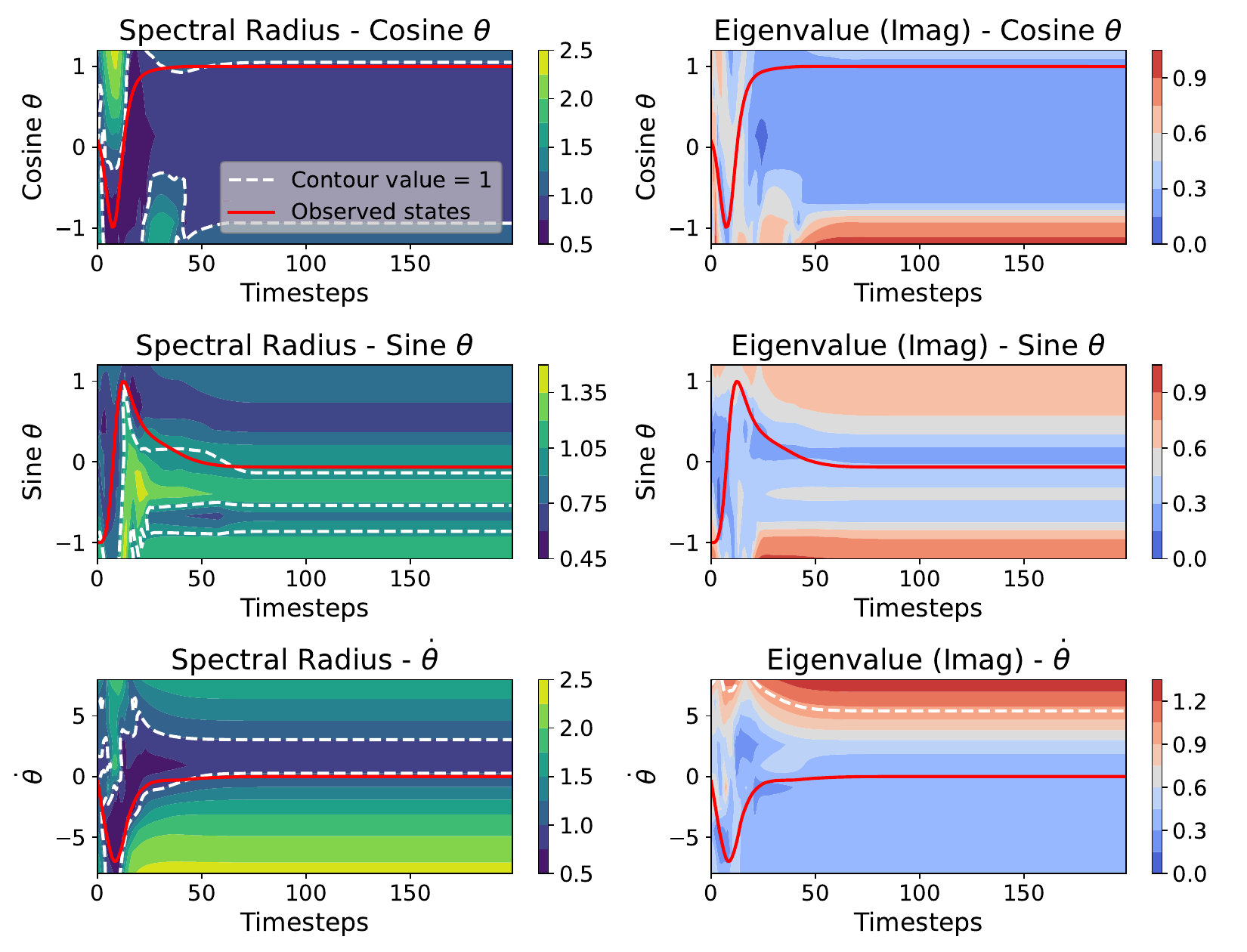}
    \hfill
    \includegraphics[width=0.49\textwidth]{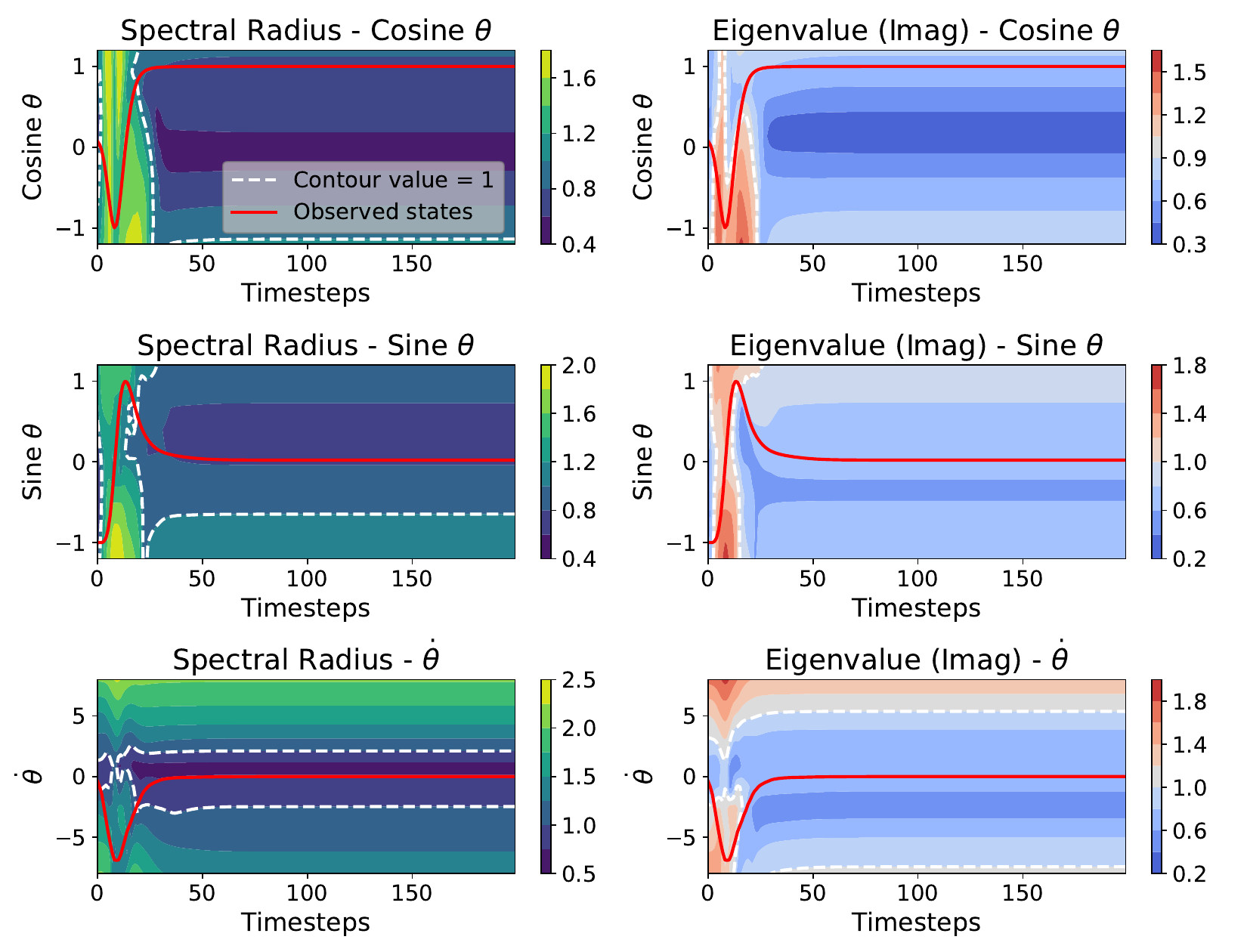}
    \caption{Effect of hidden dimensions on local stability. Contours are shown for (top-left) $h_d=3$, (top-right) $h_d=4$, (bottom-left) $h_d=8$, and (bottom-right) $h_d=16$. Consistent stability patterns across all configurations demonstrate the robustness of the analysis to latent dimension size.}
    \label{fig:pendulum_stability_hd}
\end{figure}

In Figure \ref{fig:pendulum_stability_hd}, the effect of different hidden dimensions contributing to the stability analysis is extensively demonstrated. 
Overall, the stability patterns remain consistent across different $h_d$ configurations: the trajectory initially passes through high-instability regions (yellow) before stabilizing over time.

In the first row with low $h_d$, the contour patterns are highly similar within the spectral radius and within the imaginary component plots. This consistency aligns with the main article and is particularly evident in the initial stage, where the control policy seeks local stability while crossing regions of instability.
In the second row with high $h_d$, the spectral radius contour reveals more symmetric and well-defined regions of stability, in contrast to the upper row, where local stability is only observed on the lower-value side and the upper region remains unbounded or weakly constrained. This suggests that higher latent dimensions enable more accurate local stability characterization, particularly evident in the angular velocity plot.

It is worth noting that in the higher $h_d$ cases, even after the pendulum stabilizes (beyond 50 timesteps), the imaginary component reveals regions of high values, indicating oscillatory behavior. However, these regions remain distant from the actual state of the pendulum, and the increased values are likely due to the higher-dimensional neural network introducing greater nonlinearity.

\section{Phase-Space Visualizations of \methodName{}}\label{appendix:results_structured_action}
For the pendulum problem, the state space is summarized by the angle $\theta$ and angular velocity $\dot{\theta}$ instead of the trigonometric components $\cos\theta$ and $\sin\theta$. The phase-space behavior of the learned policies is analyzed by interacting the policy with the environment, allowing for the visualization of the action structure assigned to different states.

\begin{figure}[ht!]
\begin{center}
\centerline{\includegraphics[width=0.8\columnwidth]{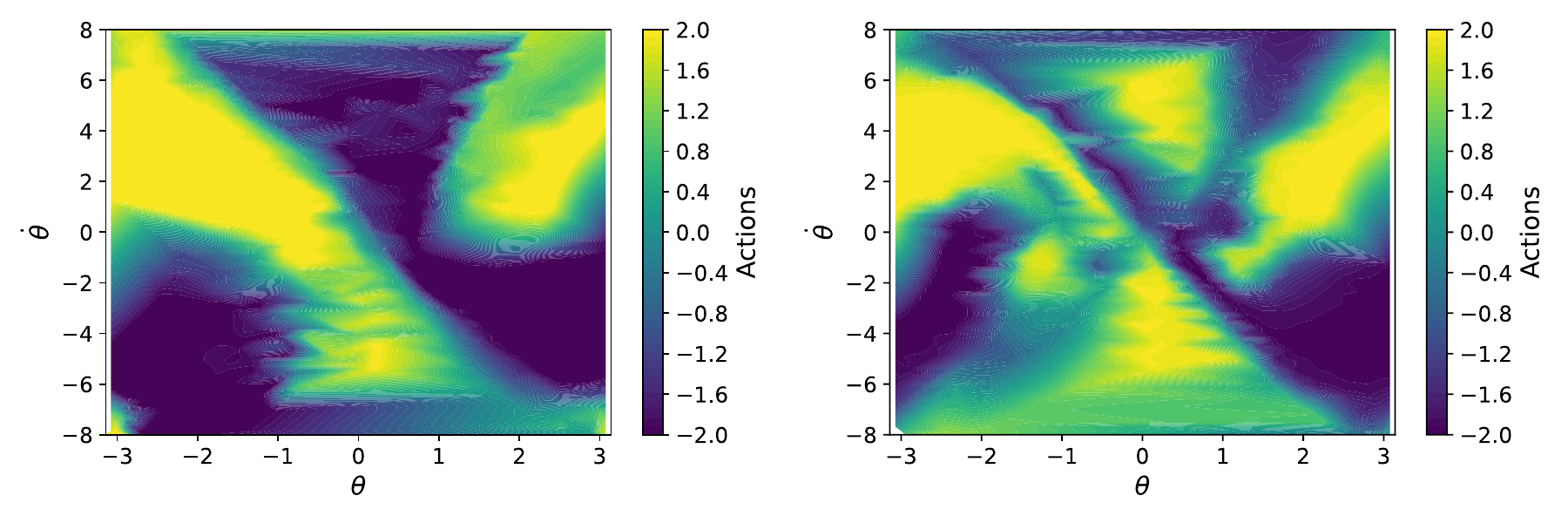}}
\caption{Action space comparison between \methodName{} (left) and SAC (right). \methodName{} exhibits a more structured action space with sharper transitions, whereas SAC displays smoother gradients across the phase space.}
\label{fig:appendix_pendulum2}
\end{center}
\end{figure}

Figure~\ref{fig:appendix_pendulum2} compares the baseline SAC \cite{rl-zoo3} with the proposed \methodName{}, both sharing similar architectures and training strategies. On the right, the SAC policy results in scattered action assignments across phase space, reflecting its probabilistic nature. In contrast, on the left, \methodName{} exhibits larger, more coherent action regions, indicating more consistent action choices for given states. This suggests that \methodName{} encodes a structured latent representation, leading to more predictable decision-making in phase space.
These slight discrepancies arise from the encoder-decoder structure in \methodName{}, which transforms raw actions into a learned latent space before applying them, thus altering the effective action dynamics. Notably, this transformation may also enhance the stability of the pretrained agent, though a full analysis of this effect is left for future work.
Like previous work \cite{weissenbacher2022koopman}, this approach may also be used to perform principled data augmentation for improved behavior of RL agents in terms of regularity.

To assess generalizability and robustness, both policies were tested in unseen pendulum environments by modifying internal physical parameters, such as gravity and rod mass. Compared to SAC, \methodName{} maintained control with up to a 40\% mass increase or a 20\% gravity increase, while SAC failed to keep the pendulum upright in all variations. This suggests that \methodName{}'s structured latent dynamics contribute to improved stability and performance across diverse conditions.  

\end{document}